\documentclass[journal]{IEEEtran}

\IEEEoverridecommandlockouts
\usepackage{graphicx}
\def\BibTeX{{\rm B\kern-.05em{\sc i\kern-.025em b}\kern-.08em
    T\kern-.1667em\lower.7ex\hbox{E}\kern-.125emX}}

\usepackage{blindtext}
\usepackage{caption}

\let\oldtwocolumn\twocolumn
\renewcommand\twocolumn[1][]{%
    \oldtwocolumn[{#1}{
    \begin{center}
    \vskip-5ex
        \centering
        \includegraphics[width=0.95\textwidth]{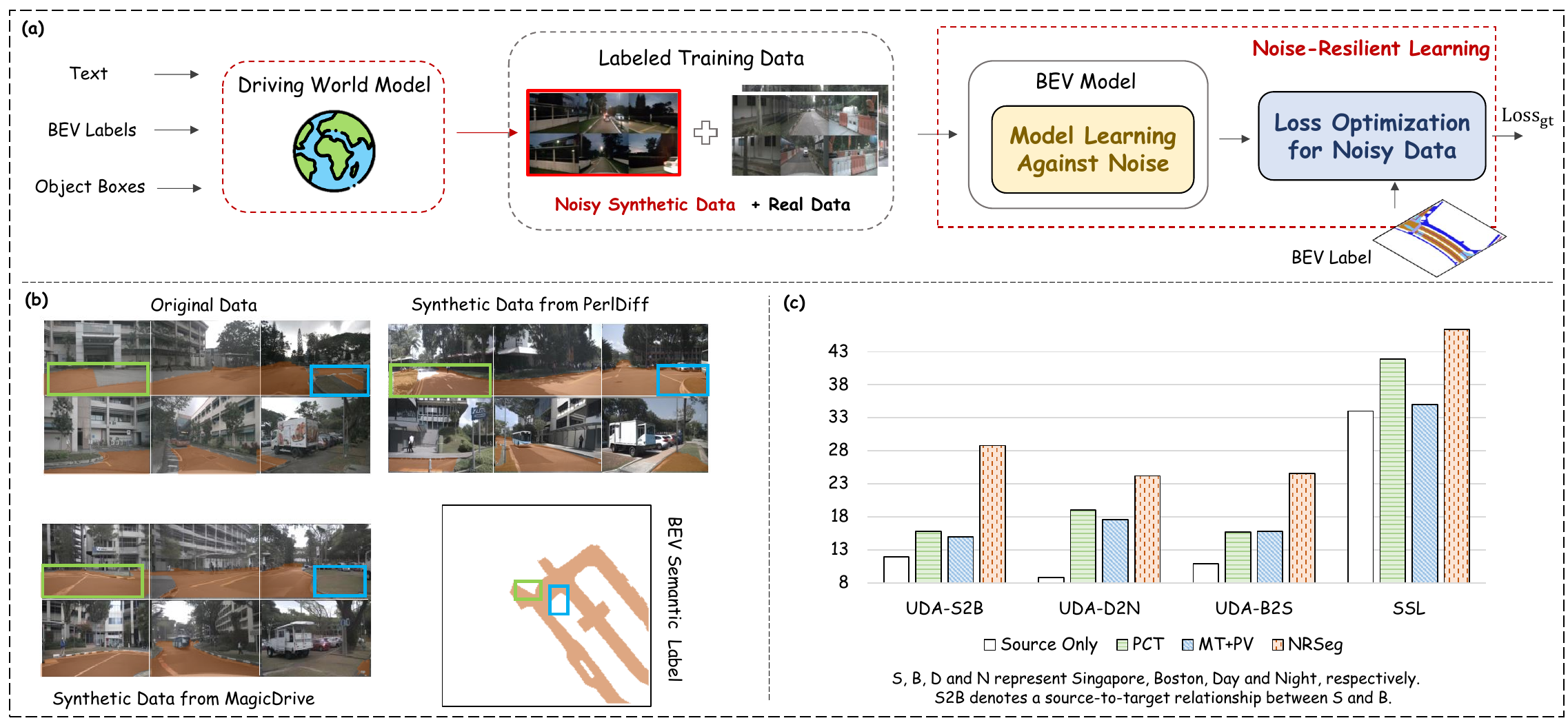}
        \captionof{figure}{\small
        Our approach is motivated by the potential of leveraging noisy synthetic data from driving world models to enhance BEV semantic segmentation. 
        (a) The pipeline of NRSeg for BEV semantic segmentation via driving world models. 
        The proposed method investigates a noise-resilient learning framework designed for handling synthetic data with inherent noise.
        (b) Visualization of data noise in world models~\cite{zhang2024perldiff,gao2023magicdrive}. 
        The generated data from different world models exhibits inconsistent road structures at identical viewpoints. 
        (c) In Unsupervised Domain Adaptation (UDA) and Semi-Supervised Learning (SSL) tasks, NRSeg outperforms the state-of-the-art BEV segmentation method PCT~\cite{pct}.  
      }
      \label{Fig.intro_new}
    \end{center}
    \vspace {-0.6em}
    }]
}

\usepackage{multirow} 
\usepackage{amsmath,amsfonts}
\usepackage{algorithmic}
\usepackage{algorithm}
\usepackage{array}
\usepackage{booktabs}
\usepackage[caption=false,font=normalsize,labelfont=sf,textfont=sf]{subfig}
\usepackage{textcomp}
\usepackage{stfloats}
\usepackage{url}
\usepackage{verbatim}
\usepackage{graphicx}
\usepackage{cite}

\usepackage{amsmath}

\usepackage[table,xcdraw]{xcolor}
\usepackage{amssymb}
\usepackage{xcolor}
\definecolor{rblue}{rgb}{0,0.5,1}
\definecolor{hollywoodcerise}{rgb}{0.96, 0.0, 0.63}
\definecolor{lasallegreen}{rgb}{0.03, 0.47, 0.19}
\definecolor{hanpurple}{rgb}{0.32, 0.09, 0.98}
\definecolor{green(pigment)}{rgb}{0.0, 0.65, 0.31}
\usepackage[pagebackref=false,breaklinks=true,colorlinks,bookmarks=false]{hyperref}
\hypersetup{colorlinks,linkcolor={red},citecolor={hanpurple},urlcolor={magenta}}  
\DeclareUnicodeCharacter{2061}{}

\newcommand{\lsy}[1]{\textcolor{black}{#1}}
\begin{document}

\title{
NRSeg: Noise-Resilient Learning for BEV Semantic Segmentation via Driving World Models
}

\author{Siyu Li$^{1,*}$, Fei Teng$^{1,*}$, Yihong Cao$^{2}$, Kailun Yang$^{1}$, Zhiyong Li$^{1}$, and Yaonan Wang$^{1}$
\thanks{This work was supported in part by the National Natural Science Foundation of China (Grant No. U21A20518, No. 61976086, and No. 62473139), in part by the Hunan Provincial Research and Development Project (Grant No. 2025QK3019), and in part by the State Key Laboratory of Autonomous Intelligent Unmanned Systems (the opening project number ZZKF2025-2-10). 
\textit{(Corresponding authors: Kailun Yang and Zhiyong Li.)}
}
\thanks{$^{1}$S. Li, F. Teng, K. Yang, Z. Li, and Y. Wang are with the School of Artificial Intelligence and Robotics and the National Engineering Research Center of Robot Visual Perception and Control Technology, Hunan University, Changsha 410082, China (email: kailun.yang@hnu.edu.cn; zhiyong.li@hnu.edu.cn).}
\thanks{$^{2}$Y. Cao is with the Key Laboratory of Big Data Research and Application for Basic Education, Hunan Normal University, Changsha 410006, China.}
\thanks{$^{*}$Equal contribution.}
}

\markboth{IEEE Transactions on Image Processing, February~2026}%
{Li~\MakeLowercase{\textit{et al.}}: NRSeg.}

\maketitle

\begin{abstract}
Birds' Eye View (BEV) semantic segmentation is an indispensable perception task in end-to-end autonomous driving systems. Unsupervised and semi-supervised learning for BEV tasks, as pivotal for real-world applications, underperform due to the homogeneous distribution of the labeled data. In this work, we explore the potential of synthetic data from driving world models to enhance the diversity of labeled data for robustifying BEV segmentation. Yet, our preliminary findings reveal that generation noise in synthetic data compromises efficient BEV model learning. To fully harness the potential of synthetic data from world models, this paper proposes NRSeg, a noise-resilient learning framework for BEV semantic segmentation. Specifically, a Perspective-Geometry Consistency Metric (PGCM) is proposed to quantitatively evaluate the guidance capability of generated data for model learning. This metric originates from the alignment measure between the perspective road mask of generated data and the mask projected from the BEV labels. Moreover, a Bi-Distribution Parallel Prediction (BiDPP) is designed to enhance the inherent robustness of the model, where the learning process is constrained through parallel prediction of multinomial and Dirichlet distributions. The former efficiently predicts semantic probabilities, whereas the latter adopts evidential deep learning to realize uncertainty quantification. Furthermore, a Hierarchical Local Semantic Exclusion (HLSE) module is designed to address the non-mutual exclusivity inherent in BEV semantic segmentation tasks.
The proposed framework is evaluated on BEV semantic segmentation using data generated by multiple world models, with comprehensive testing conducted on the public nuScenes dataset under unsupervised and semi-supervised settings. Experimental results demonstrate that NRSeg achieves state-of-the-art performance, yielding the highest improvements in mIoU of $13.8\%$ and $11.4\%$ in unsupervised and semi-supervised BEV segmentation tasks, respectively. The source code will be made publicly available at \url{https://github.com/lynn-yu/NRSeg}.

\end{abstract}

\begin{IEEEkeywords}
Bird's-eye-view understanding, driving world model, 
unsupervised domain adaptation, 
evidential deep learning. 
\end{IEEEkeywords}

\section{Introduction}
Bird's Eye View (BEV) semantic segmentation has emerged as a critical task in the autonomous driving community~\cite{bevsurvey,mapsurvey}. 
While the accuracy of BEV semantic perception continues to improve with ongoing research~\cite{LSS,pmapnet}, the precise segmentation of unconstrained driving surroundings remains a significant challenge.

To improve BEV understanding in real-world scenes, Unsupervised Domain Adaptation (UDA) and Semi-Supervised Learning (SSL) have become prevalent approaches for deploying perception tasks in unlabeled environments.
In BEV semantic segmentation, numerous studies have explored these two methodologies. 
Specifically, DualCross~\cite{man2023dualcross} employed adversarial learning to achieve UDA, while PCT~\cite{pct} introduced perspective pseudo-label supervision for both UDA and SSL. SemiBEV~\cite{semibev}, designed for single-view tasks, enhanced SSL efficiency through a spatially-rotated data augmentation strategy.
Although these methods have improved the accuracy of UDA and SSL in BEV perception, they fundamentally rely on existing annotated data, thereby achieving only limited improvements in learning capability.
\lsy{Furthermore, since the BEV semantic segmentation task is highly data-hungry, with its performance heavily dependent on the scale of available labeled data, enhancing labeled training data can significantly improve model performance~\cite{much}.} 
However, the annotation process of BEV labels remains highly labor-intensive.
The rapid advancement of generative models~\cite{freemask, datasetdiffusion, trainingfree} has attracted our attention, as these models can synthesize highly diverse and photorealistic images from labels.
Thus, recent advances in driving world models~\cite{bevcontrol,gao2023magicdrive,zhang2024perldiff}, which generate multi-view street images from BEV semantic labels, vehicle bounding boxes, and text prompts, have raised an important yet underexplored question: 
Does incorporating diversely distributed synthetic images with BEV labels improve BEV semantic perception performance?

In this work, we look into this conjecture and aim to unveil the potential of driving world models. 
On one hand, diversified synthetic data has the potential to enhance the learning capability of BEV semantic segmentation models. On the other hand, world models serve as abstract environmental representation systems for predicting future states. Utilizing generated synthetic data to train BEV models facilitates a deeper understanding of intrinsic environmental representations.
This represents the first systematic study examining the efficacy of synthetic data for BEV semantic segmentation. 
Interestingly, via exploratory experiments, while the synthetic data can effectively enhance BEV perception capacity, a noticeable performance gap persists compared to directly incorporating the original data of street-view scenes.
Following our observations as shown in Fig.~\ref{Fig.intro_new}(b), while current world models have strong reasoning capabilities, their interpretation of control signals is imperfect, causing structural drift in the synthetic images and resulting in misalignment with the true road geometry. 
Therefore, directly using these labels to guide the learning of synthetic data can adversely affect the accurate understanding of the view transformer between the perspective view and BEV. To unlock the potential of data generated by world models, noise-resilient learning of synthetic data under limited supervision has become a key factor for improving performance in BEV perception tasks.

To enhance the noise robustness of BEV models, this work proposes a novel paradigm, NRSeg, a noise-resilient learning framework for BEV semantic segmentation, as shown in Fig.~\ref{Fig.intro_new}(a).
The proposed method simultaneously explores optimized guidance for synthetic data of varying quality and robust learning of the model itself against disturbances.
A Perspective-Geometry Consistency Metric (PGCM) is designed to quantitatively assess the contribution of synthetic data to model learning.
From the perspective view, the contribution scores of synthetic data are quantified by evaluating the similarity between road masks in synthetic images and their BEV back-projected counterparts.
To alleviate the effect of generation noise in synthetic data, this score adaptively encourages the model to focus on non-labeled regions of synthetic data by guiding the loss optimization direction.
Furthermore, the model's robustness against disturbances is enhanced through dual-distribution constraints on predictions in the Bi-Distribution Parallel Prediction (BiDPP) module.
While the multinomial distribution directly predicts semantic probabilities of a single pixel, the Dirichlet distribution is employed to model the uncertainty relationships based on Evidential Deep Learning (EDL)~\cite{evidential,edl2uda-2}.
The premise of EDL requires mutually exclusive semantic categories, which cannot be satisfied in BEV tasks.
To effectively adapt EDL for BEV semantic segmentation tasks, the Hierarchical Local Semantic Exclusion (HLSE) module is designed, which partitions locally mutually exclusive semantic categories into unit clusters and achieves uncertainty modeling through hierarchical fusion of their independent learning processes.

We evaluate NRSeg under unsupervised domain adaptation and semi-supervised learning tasks assembled by the nuScenes datasets~\cite{nus}. 
Extensive experiments show that our model reaches state-of-the-art performance in BEV semantic segmentation with significant improvements of $13.8\%$ and $11.4\%$ in unsupervised and semi-supervised tasks.
The main contributions delivered in this work are summarized as follows:
\begin{itemize}
\item This paper first proposes NRSeg, a noise-resilient learning framework for BEV semantic segmentation, aiming to unleash the full potential of synthetic data via driving world models for enhancing BEV segmentation models.
\item A perspective-geometry consistency metric is proposed to evaluate the contribution of synthetic data to model learning, thereby guiding the optimization direction with quantitative control.
\item A bi-distribution parallel prediction module employs parallel prediction of both semantically independent and mutually exclusive distributions, while designing a hierarchical semantic exclusivity module to enable uncertainty modeling under globally non-exclusive conditions.
\item A comprehensive variety of experiments demonstrates that NRSeg achieves remarkable performance in both unsupervised domain adaptation and semi-supervised learning tasks, significantly advancing the capabilities of BEV semantic segmentation models.
\end{itemize}

\section{Related Work}
\label{related}
\subsection{BEV Semantic Segmentation}
BEV semantic mapping performs semantic segmentation of static objects in road environments from a Bird's Eye View (BEV). The fundamental challenge of BEV semantic segmentation lies in inferring 3D information from 2D inputs. Depending on the approach to view transformation, this research currently encompasses three distinct methodological categories.
One dominant approach, exemplified by the LSS framework~\cite{LSS,bevpoolv2}, achieved spatial transformation through 2D pixel depth prediction. The alternative paradigm adaptively learned spatial feature representations by projecting 3D sampling points onto 2D image planes~\cite{BEVFormer,BEVFormerv2}.
Furthermore, some approaches~\cite{hdmapnet,pmapnet} leveraged the prior knowledge that static road objects predominantly lie on the ground plane, utilizing this prior information to facilitate view transformation through Inverse Perspective Mapping (IPM).
While these view transformation methods demonstrate efficient adaptation to diverse road environments, their performance significantly degrades in complex, congested scenarios due to the sparsity of 2D observations~\cite{shan2025stability}. To address this, StreamMapNet~\cite{streammapnet} introduced a streaming temporal strategy~\cite{wang2023streampetr,lin2023sparse4d} that compensated for single-frame sparsity by aggregating long-term historical information, improving mapping accuracy.

However, BEV semantic segmentation models exhibit strong dependence on data distribution. When applied to unlabeled tasks across diverse environments, the models experience significant performance degradation~\cite{semibev}.  
To address model generalization limitations, recent work~\cite{genmapping} has developed parameter-decoupled view transformation methods that isolate geometric reasoning from the learning process, thereby enhancing cross-dataset performance.  
For mitigating inter-domain disparities, unsupervised domain adaptation presents a more effective solution.
DualCross~\cite{man2023dualcross} introduced adversarial learning into BEV semantic segmentation, aligning feature distributions across domains to enhance cross-domain performance.
PCT~\cite{pct} adopted a mean teacher framework for domain adaptation, leveraging perspective-view semantic labels to enhance the cross-domain learning capability.

Although these methods demonstrate competitive performance across domains, they have not yet explored directly enhancing the learning capability of the model by increasing the diversity of data distributions within the source domain.  
Benefiting from the rapid advancement of generative models~\cite{yang2023diffusion,controlnet}, it is now feasible to utilize simple yet effective information to synthesize diverse data. 
We aim to explore the potential of synthetic data from driving world models for robustifying BEV semantic segmentation.

\subsection{Semantic Segmentation with Synthetic Data}
Given the labor-intensive nature of pixel-level annotation and the maturity of contemporary image generation models, numerous studies now directly employ generative models to synthesize both images and their corresponding labels. 
DiffuMask~\cite{wu2023diffumask} represented a text-conditioned framework that simultaneously generated images and their corresponding semantic masks, which exploited the potential of the cross-attention map between text and image. 
Building upon this, Dataset Diffusion~\cite{datasetdiffusion} incorporated class prompts to enhance both instance diversity and generation accuracy.
Distinct from these text-to-image-label generation approaches, alternative methods~\cite{freestyle,freemask} leveraged both textual descriptions and semantic labels to synthesize diversified images. 
To establish precise spatial correspondence between text and semantic labels, FreestyleNet~\cite{freestyle} introduced a rectified cross-attention module to address this alignment challenge. 
FreeMask~\cite{freemask} proposed an active learning paradigm to mitigate quality degradation in image generation caused by occlusion issues.
Although synthetic data significantly enhances semantic segmentation performance, the presence of low-quality samples provides erroneous guidance to the model. 
Thus, the work of~\cite{trainingfree} proposed a training-free CLIP-based image selection strategy that filters out low-quality samples from the training corpus, thereby improving segmentation quality.

These synthetic data generation and optimized methods primarily target single-view perspective image segmentation tasks. In contrast, BEV tasks inherently involve multi-view images and the view transformer, where world models currently represent a potential approach.

\subsection{Driving World Models for Street-View Generation}
World models can establish an internal representation framework that learns 2D projections from 3D structures, facilitating environmental comprehension and future state prediction~\cite{worldmodelsurvey}. 
The BEV model trained by synthetic data can analyze these 2D images to further understand the underlying internal representations, thereby forming a closed-loop system.

Existing world model research~\cite {hafner2019dream,mile,sem2} primarily focused on virtual environments, employing reinforcement learning or deep learning methodologies. However, for autonomous driving applications, the complexity of real-world environments necessitates training with more realistic and representative environmental data.
Consequently, DriveDreamer~\cite{DriveDreamer} proposed the first world model fully derived from real-world driving scenarios. Leveraging BEV maps and object bounding boxes, it achieved controllable driving video generation while simultaneously predicting behavioral decisions.
To enhance generated image quality, BEVControl~\cite{bevcontrol} implemented a two-stage world model architecture based on diffusion models. 
MagicDrive~\cite{gao2023magicdrive} further incorporated 3D object boxes as an additional control signal and diverse static road elements into the map representation, including drivable areas, lane markings, crosswalks, stop lines, and parking slots. 
It significantly enhanced the realism of generated images by closely mirroring real-world environmental complexity.
PerlDiff~\cite{zhang2024perldiff} similarly employed this diversified BEV map control strategy. Additionally, it projects the BEV map onto perspective views to create perspective road masks as supplementary control conditions. These geometric priors enable more accurate view transformation learning, thereby achieving finer-grained control over street scene generation.
Furthermore, beyond employing 2D maps as control conditions, contemporary world models incorporate multi-modal inputs~\cite{HoloDrive}, occupancy states~\cite{zheng2024occworld}, and temporal sequences~\cite{MagicDrive-V2} as auxiliary conditions to enhance their research framework.
Although synthetic data generated via driving world models offers significant advantages for BEV perception tasks, geometric discrepancies between the generated images and corresponding BEV label representations in complex environments limit their optimal application.
To address this fundamental challenge, our work pioneers a novel noise-resilient learning framework that leverages correlation scoring and uncertainty analysis to enhance model robustness significantly.

\begin{figure*}[tb]
      \centering
      \includegraphics[scale=0.46]{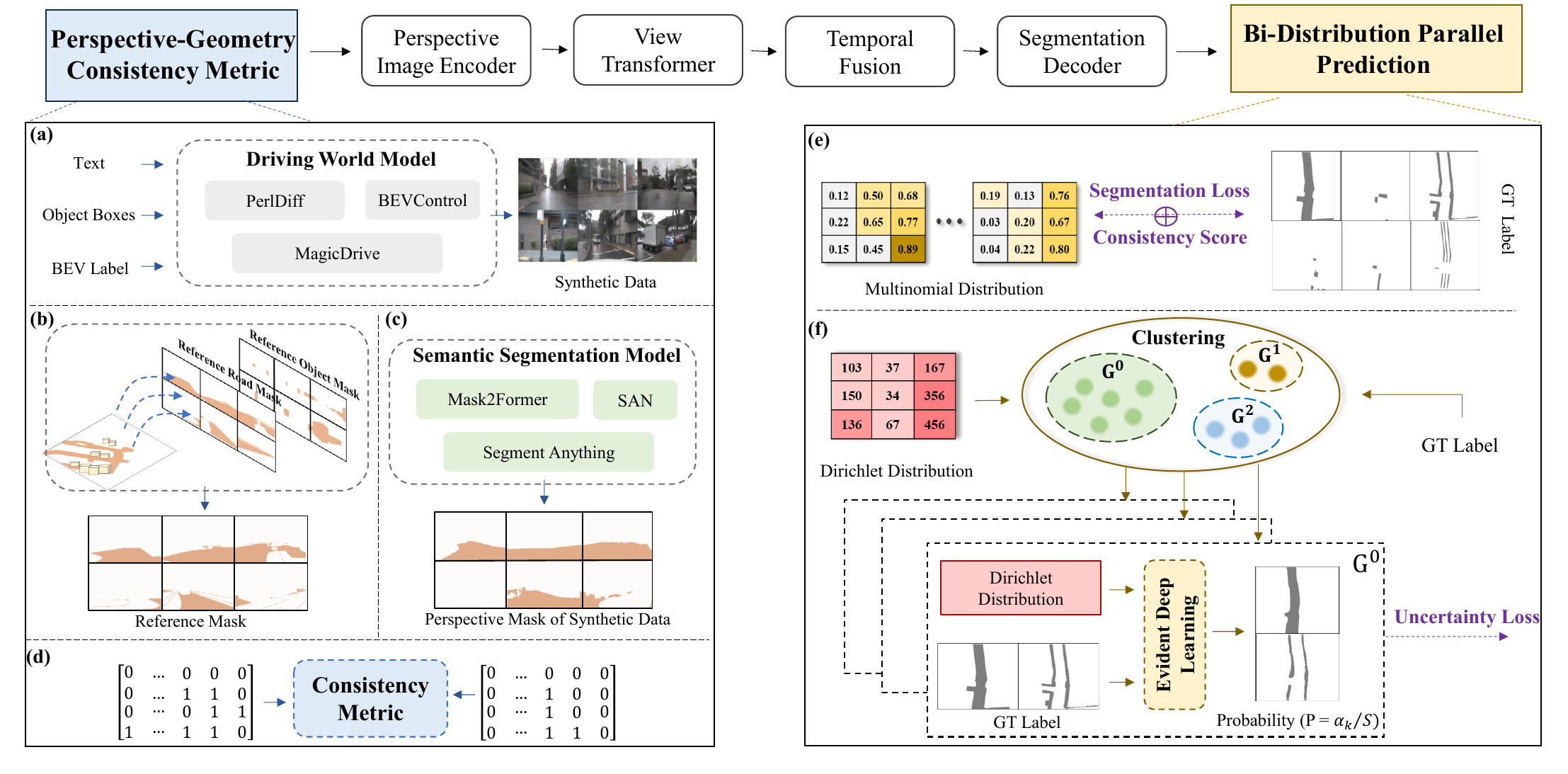}
      \vspace{-1.0em}
      \caption{Illustration of the proposed NRSeg framework for BEV semantic segmentation. 
      It includes a Perspective-Geometry Consistency Metric (PGCM) and a Bi-Distribution Parallel Prediction (BiDPP) module. 
      PGCM includes (a) synthetic data jointly are generated by different world models using BEV labels, bounding boxes, and text; (b) reference masks are generated by back-projecting BEV semantic labels; (c) synthetic masks are produced using a general semantic segmentation model; (d) consistency scores of synthetic data are learned based on mask.
      BiDPP innovatively co-learns multinomial and Dirichlet distributions where (e) represents the multinomial distribution for semantic predictions, while introducing a consistency score derived from synthetic data noise analysis into the segmentation loss; (f) innovatively designs the Hierarchical Local Semantic Exclusion (HLSE) module to fully leverage evidential deep learning theory for fine-grained uncertainty quantification. 
      }
      \label{Fig.frame}
      \vspace{-1.5em}
\end{figure*}
\subsection{Evidential Deep Learning}
Evidential Deep Learning (EDL) is a probabilistic framework that enhances prediction accuracy through explicit uncertainty quantification~\cite{evidential}.
This uncertainty-quantified learning paradigm has been widely adopted in semantic segmentation~\cite{edl2panaseg} and action recognition~\cite{edl2open} tasks, demonstrating significant improvements in robustness and interoperability. 
Moreover, these works~\cite{edl2uda-1,edl2uda-2} proved effective for unsupervised domain adaptation tasks by identifying out-of-distribution samples, thereby enhancing model robustness. Building upon this, the work of~\cite{edl2uda-2} further developed an advanced uncertainty estimation method to facilitate the fusion of multi-source domain knowledge.
The applications in BEV tasks remain relatively scarce. The work~\cite{uncbev} systematically analyzed the impact of various uncertainty methods on model robustness, demonstrating the potential of evidential learning. 
The work~\cite{edl2BEV} further corroborated these findings, showing that uncertainty prediction through evidential learning can enhance object detection accuracy in BEV space.
Since EDL fundamentally requires learning mutually exclusive categories, existing BEV studies employing this approach have been limited to single-category segmentation tasks. This constraint arises because not all static object categories in BEV space exhibit inherent mutual exclusivity, \textit{e.g.}, drivable areas and lane markings can coexist spatially.
To fully unleash the potential of EDL in BEV semantic segmentation, we propose a Hierarchical Local Semantic Exclusion (HLSE) module that guides global uncertainty learning through mutual semantic category interactions at local scales.

\section{Method}
In this section, we first introduce the framework of the proposed NRSeg for noise-resilient BEV semantic segmentation in Sec.~\ref{frame}. 
Then, facing the noise in synthetic data generated by world models, the Perspective-Geometry Consistency Metric (PGCM) module (Sec.~\ref{PGCM}) is designed to optimize the guided learning of noisy data, whereas the Bi-Distribution Parallel Learning (BiDPP) module (Sec.~\ref{BiDPP}) is constructed to enhance the robustness of the model.

\subsection{Framework of the NRSeg Model}
\label{frame}
Typically, a BEV semantic segmentation model consists of three basic modules: a perspective image encoder, a view transformation module, and a BEV segmentation decoder. The multi-view images $I_n$ are encoded with the perspective image encoder to obtain image features $F_i$. 
Furthermore, these features can be utilized to generate perspective semantic segmentation predictions $P_v$, where the perspective mask ground truth $L_v$ supervises the perspective-view encoding module by the segmentation loss:
\begin{equation}
    Loss_{pv} = \text{SegLoss}(P_v,L_v).
    \vspace{-0.3em}
\end{equation}
Then, the view transformer module translates these features into the BEV space $F_b$. Considering the sparsity of visual information, current works~\cite{BEVFormer,streammapnet} have integrated temporal fusion modules into the models through a stacking or streaming strategy. They achieve multi-frame BEV feature $F_b^{t}$ fusion through either GRU-based sequential modeling~\cite{gru} or attention mechanisms~\cite{attention}. 
Compared to the computational redundancy of stacked temporal fusion strategies, the streaming approach offers greater lightweight efficiency and faster processing. The proposed work adopts the streaming fusion strategy to realize more accurate segmentation. 
Specifically, the temporal fusion module $\text{TFu}$ takes the previous frame $F_b^{t-1}$ after alignment as latent state input to guide fusion with the current frame $F_b^{t}$, thereby generating more comprehensive BEV features $F_b^{t'}$.
\begin{equation}
\vspace{-0.3em}
    F_b^{t'} = \text{TFu}(\text{Ailgh}(F_b^{t-1}),F_b^{t}).
\end{equation}
Additionally, a temporal alignment loss $Loss_a$ is incorporated as an auxiliary constraint. During vehicle motion, consecutive frames exhibit viewpoint discrepancies. The temporal constraint is applied only to overlapping field-of-view regions across frames. The overlapping field-of-view mask can be calculated as:
\begin{equation}
    m_c = \text{Aligh}(\begin{bmatrix}
  1  &\dots  &1 \\
  1  &\dots  &1 \\
  1   &\dots  &1 
\end{bmatrix}_{h_f*w_f}),
\end{equation}
where $h_f$ and $w_f$ are the same dimensions of BEV features. The temporal alignment loss $Loss_a$ is obtained:
\begin{equation}
    Loss_{tr} = |m_c * (\text{Ailgh}(F_b^{t-1}) - F_b^{t})|.
\end{equation}
Moreover, $N$ is used to contain a shared BEV encoder and two segmentation detectors: a multinomial distribution $D_m^k$ and a Dirichlet distribution $D_d^k$ for predictions. 
\begin{equation}
\vspace{-0.5em}
    D_m^k,D_d^k = N(F_b^{t'}),
\end{equation}
where $k$ is the number of semantic classes. The semantic probability predictions can be obtained by $P_m^k= \text{Sigmoid}(D_m^k)$.
Typically, the BEV semantic segmentation models are supervised by semantic labels $L_b$, which guide model optimization by constraining the predicted semantic probabilities $P_m^k$. \lsy{Following the semi-supervised work~\cite{semibev}, the DICE loss is adopted:}
\begin{equation}
\label{lossgt}
    Loss_{gt} = 1-\frac{\sum_k\sum_{h=0}^{H_b}\sum_{w=0}^{W_b}2*P_m^k*L_b^k}{\sum_k\sum_{h=0}^{H_b}\sum_{w=0}^{W_b}(P_m^k+L_b^k+\sigma)} .
\end{equation}

In contemporary BEV perception datasets, perspective images and semantic labels are precisely aligned. 
Yet, due to generation errors in world models, synthetic data and semantic labels cannot maintain perfect correspondence.
As shown in Fig.~\ref{Fig.frame}, to mitigate the misleading effects of such misalignment, this paper proposes the Perspective-Geometry Consistency Metric (PGCM) and Bi-Distribution Parallel Learning (BiDPP) to strengthen noise-resilient learning and model robustness in BEV semantic segmentation.

\subsection{Perspective-Geometry Consistency Metric}
\label{PGCM}
Although world models employ BEV semantic maps to control image generation, significant generation noise persists in complex road environments due to insufficient understanding of control information, resulting in imperfect alignment with BEV semantic labels, as shown in Fig.~\ref{Fig.intro_new}. 
While incorporating labeled synthetic data into training can enhance model learning capacity, enforcing strict fitting between misaligned data and labels may lead to erroneous learning trajectories.
To ensure effective learning from synthetic data, in this section, we design a perspective-geometry consistency metric to evaluate data accuracy, which serves as a constraint factor to guide the loss optimization process for improving the utilization of the generated synthetic data.

First, the BEV semantic label and 3D object boxes are projected onto the perspective view, maintaining consistency with the process described in works~\cite{maptrv2,genmapping,zhang2024perldiff}. 
Assuming the BEV label lies on the ground plane, the height is $h=0$.
With this prior knowledge, each cell $(X,Y)$ of the BEV label $M_b$ can be projected into perspective views $(u,v)$:
\begin{equation}
\label{eq1}
\vspace{-0.3em}
    [u,v,1]^\text{T} = \text{Proj}_{\text{b2p}}(T_{\text{ex}}, T_{\text{in}}, [X,Y,h]^\text{T}),
\end{equation}
where $T_{ex}$ and $T_{in}$ is the camera parameters. $Proj_{b2p}(\cdot)$ is the conversion from BEV coordinates to perspective coordinates. 
Then the projected road mask of perspective view $M_{o}$ can be obtained. 
The 3D object boxes are directly back-projected onto the perspective image through Eq.~\ref{eq1} to obtain the projected object mask $M_{c}$.
The perspective view is a 2D representation, inherently suffering from occlusion issues due to viewpoint limitations. Since targets like vehicles exist above the ground plane, they occlude the road surface during perspective imaging. Consequently, the final reference perspective mask should adhere to the following criteria:
\begin{equation}
\vspace{-0.5em}
    M_r = M_o \cap \overline{M_c},
\end{equation}
where $\overline{M_c}$ is the negation of $M_c$.
For synthetic images, perspective masks $M_s$ can be produced by pre-trained semantic segmentation models such as Mask2Former~\cite{mask2former} and SAN~\cite{san}.

To achieve accurate correspondence between synthetic data and BEV labels, the reference and synthetic perspective masks should inherently maintain geometric consistency. 
In other words, any misalignment between them indicates inherent discrepancies between the synthetic map and BEV semantic labels.
Therefore, the alignment error between the reference perspective masks $M_r$ and the synthetic perspective masks $M_s$ is used to quantify the BEV accuracy of synthetic images. 
Considering that Intersection over Union (IoU) is a well-established metric for mask alignment evaluation, we adopt it as the primary measure here. The score of each frame oriented from synthetic data can be calculated by:
\begin{equation}
\vspace{-0.5em}
    \textbf{R} = \frac{1}{n}\sum_{i=0}^n\text{IoU}(M_{r}^i,M_{s}^i),
\end{equation}
where $n$ is the number of perspective images in each frame. 
Then, for misaligned synthetic data samples and labels, this consistency score is incorporated as a weighting factor to guide the loss optimization direction in deep learning model training.

\lsy{Synthetic data generated under BEV label control maintains overall structural consistency with the labels, with noise primarily manifesting as subtle structural shifts. Therefore, this paper selects the DICE loss~\cite{dice}, which favors holistic structural learning and exhibits high tolerance for minor errors.}
The DICE loss is defined as:
\begin{equation}
\label{dice}
    loss_{gt} = 1-\frac{\sum_{h=0}^{H}\sum_{w=0}^{W}2*P*L}{\sum_{h=0}^{H}\sum_{w=0}^{W}(P+L+\sigma)} ,
\end{equation}
where $H$ and $W$ are the shapes of labels. $P$ is the prediction probabilities and $L$ is the labels.
Since the labels follow a binary distribution, Eq.~\ref{dice} can be simplified as:
\begin{equation}
\label{loss_dice}
\vspace{-0.5em}
    loss_{gt} = 1-\frac{2*P_{+}}{P_{+} + P_{-}+N_{+}+\sigma},
\end{equation}
where $P_{+}$ and $P_{-}$ are the summed predicted probabilities for labeled ($L=1$) and non-labeled ($L=0$) regions, respectively. $N_+$ is the number of pixels in the labeled region $P_{+}$. Noted that this equality holds forever, $P_+ \le N_+$.  $\sigma$ is the smooth value.
The optimization objective of this loss is as follows:
\begin{equation}
\label{dice-1}
\vspace{-0.5em}
    \min_{p+,p_-}1-\frac{2*P_{+}}{P_{+} + P_{-}+N_{+}+\sigma}.
\end{equation}
Intuitively, this requires the numerator and denominator of the right-hand term in Eq.~\ref{dice-1} to be equal. In other words, this enforces two conditions: (i) $P_+ \approx N_+$, and (ii) $P_- \approx 0$. 

However, when synthetic sample predictions are not perfectly label-aligned, the two conditions are not necessarily true, where $P_+ < N_+$, and $P_- > 0$ are possible. 
If Eq.~\ref{dice-1} is enforced, it causes overfitting to incorrect labels and leads to erroneous learning directions. 
Therefore, to facilitate learning of non-labeled regions in synthetic samples ($P_-> 0$), we modify the loss function by incorporating a consistency score $\textbf{R}$ weighting factor into its denominator term: 
\begin{equation}
\label{dice-our}
\vspace{-0.5em}
    loss_{gt} = 1-\frac{2*P_{+}}{\textbf{R}*(P_{+} + P_{-}+N_{+}+\sigma)}.
\end{equation} 
Similarly, the optimization objective of this loss requires the equality of the numerator and denominator of the right-hand term in Eq.~\ref{dice-our}. 
\begin{equation}
\label{dice-2}
\vspace{-0.5em}
    2*P_{+} \approx \textbf{R}*(P_{+} + P_{-}+N_{+}+\sigma).
\end{equation}
After mathematical transformation, this approximate equality can be obtained:
\begin{equation}
\label{eq_p_}
\vspace{-0.5em}
    P_{-} \approx \frac{(2-\textbf{R})}{\textbf{R}}{\times}P_+ - N_+ - \sigma.
\end{equation}
$P_-> 0$ can be achieved when the process of optimization has $P_+ \in (N_+*\textbf{R}/(2-\textbf{R}), N_+]$. The condition of $\textbf{R}\in(0,1)$, satisfying the inequality of $\textbf{R}/(2-\textbf{R}) < 1$, ensures that the optimization process can hold.   
Additionally, this condition ensures the validity of $P_+ < N_+$.

Moreover, the proposed formulation reduces to Eq.~\ref{loss_dice} under the limit condition $\textbf{R}\xrightarrow{}1$, which implies alignment between samples and labels. 
Ultimately, Eq.~\ref{lossgt} can be optimized:
\begin{equation}
    Loss_{gt} = 1-\frac{\sum_k\sum_{h=0}^{H_b}\sum_{w=0}^{W_b}2*P_m^k*L_b^k}{\textbf{R} \cdot\sum_k\sum_{h=0}^{H_b}\sum_{w=0}^{W_b}(P_m^k+L_b^k+\sigma)}.
\end{equation}
It ensures guided learning with accurate labels while flexibly incorporating noisy synthetic data to facilitate the learning process.

\subsection{Bi-Distribution Parallel Learning}
\label{BiDPP}
In the BEV semantic segmentation task, the efficient multinomial distribution is typically adopted to predict semantic probabilities, where the loss function constrains the learning process. 
This constraint fundamentally relies on the reliability of the multinomial distribution assumption. 
In practice, however, the multinomial distribution exhibits inherent uncertainty when handling ambiguous categories, leading to potentially unreliable predictions.
Specifically, the synthetic data referenced in this study inherently contains noise, which significantly increases the likelihood of unreliable predictions.
Therefore, this section additionally introduces the Dirichlet distribution with Evidential Deep Learning (EDL) to further learn uncertainty. 
\lsy{Consequently, it forms a parallel learning module together with the multinomial distribution, ensuring the robustness of the BEV semantic segmentation task.}

Rooted in Dempster-Shafer Theory (DST) and Subjective Logic, EDL formalizes belief assignments through Dirichlet distributions, enabling principled uncertainty quantification~\cite{evidential}. 
Specifically, to learn a parametric representation of belief mass from the prediction, the Dirichlet distribution is employed as the prior  for parameterization as $[\alpha_{0},...,\alpha_{k}]$:
\begin{equation}
    \alpha_{k} = e_k+1 ,
\end{equation}
where $e_k=\text{ReLU}(D_d^{k})$ is the evidence information, which is derived from the BEV semantic segmentation prediction $D_d^{k}$.
Then, the parameterized distribution represents the density of probabilistic predictions $P_e$:
\begin{equation}
    D(P_e^{k}|{\mathbf{\alpha}}) = \left\{\begin{array}{ll}
    \frac{1}{B(\mathbf{\alpha})}{\textstyle \prod_{i=1}^{k}}p_e^{\alpha_i-1} , &P_e^{k}\in S_k  \\
	0 &otherwise  \\
    \end{array}\right\},
\end{equation}
where $S_k$ is the $k$-dimensional unit simplex. 
$B(\alpha)$ is a $k$-dimensional polynomial beta function.
Theoretically, the expected probability for each BEV grid belonging to a certain semantic class corresponds to the mean of the Dirichlet distribution:
\begin{equation}
\label{edl}
    P_e^{0} = \alpha_{0}/S,
\end{equation}
\begin{equation}
\label{edl-un}
    u = k/S,
\end{equation}
where $S=\sum_{k} \alpha_k$. 
However,  this theory requires that the semantic attributes of each pixel be mutually exclusive. Assuming two semantic attributes are non-exclusive, $\alpha_1\in(0,+\infty)$ and $\alpha_2\in(0,+\infty)$, their Dirichlet distribution means are:
\begin{equation}
   P_e^{1} = \alpha_{1}/(\alpha_{1}+\alpha_{2}),\\
   P_e^{2} = \alpha_{2}/(\alpha_{1}+\alpha_{2}).
\end{equation}

\begin{algorithm}[tb]  
  \caption{ Hierarchical Local Semantic Exclusion:} 
  \label{code}  
   \textbf{Input:} \\
      $D_p^{k}$: The Dirichlet distribution;\\
      $L$: The label of BEV semantic segmentation; \\
      $G$: Semantically exclusive local clusters.\\
    \textbf{Output:} 
       $Loss_{e}$ 
    \begin{algorithmic}[1]
    \STATE def KLLoss ($\alpha$):\\ 
    \COMMENT{The loss of KL divergence term, as described in Eq.~\ref{klloss}}
    \STATE $\alpha_k = \text{ReLU}(D_d^{k})+1$
    \STATE $\alpha = [\alpha_0, \dots, \alpha_k]$
    \FOR{$m = 0,\dots, len(G)$} 
        \STATE  $\alpha_m = []$, $ L_m = []$
        \FOR{$C_n = 0,\dots, len(G_m)$}%
            \STATE $\alpha_m.append(\alpha[G_m[C_n]])$%
            \STATE $L_m.append(L[G_m[C_n]])$
        \ENDFOR
        \STATE $S = \text{SUM}(\alpha_m,\text{dim}=0)$%
        \STATE $P_{inv}=S/\alpha_m$%
        \STATE $Loss_e^{G_m}$ can be obtained by Eq.~\ref{loss_edl}
        \STATE $loss_{kl}^{G_m} = \text{KLloss}(\alpha_m)$
        \STATE $Loss_e = Loss_e + Loss_e^{G_m} + \lambda_k \times loss_{kl}^{G_m}$ 
    \ENDFOR
   \end{algorithmic}
\end{algorithm}

It can be observed that the means of both distributions $P_e^{1}$ and $P_e^{2}$ cannot reach the expected value of $1$, indicating that neither class can be accurately identified.
Obviously, BEV semantic segmentation cannot directly satisfy this prior condition. 
For instance, in BEV semantic segmentation tasks, a single pixel may simultaneously belong to both drivable areas and lane markings.

Therefore, to fully utilize EDL for uncertainty modeling, the issue of semantic mutual exclusivity must be addressed. 
We conduct a systematic analysis of semantic label space relationships. 
Although BEV semantic segmentation globally violates the single-pixel mutual exclusivity principle, certain local semantic categories can satisfy this requirement when analyzed regionally. 
Consequently, we propose the Hierarchical Local Semantic Exclusion (HLSE) module to implement these mutually exclusive learning relationships.
Furthermore, this hierarchical local mutual exclusivity can construct more comprehensive relationships for learning uncertainty.

Specifically, semantic classes are first hierarchically grouped based on their inherent properties, forming multi-level local clusters. \lsy{For instance, in BEV semantic segmentation tasks, although drivable areas and lane markings are spatially non-mutually exclusive in terms of semantics, drivable areas and walkways are semantically mutually exclusive. Therefore, drivable areas and walkways can form a local cluster, $G_m$.
}
\begin{gather}
    G =\{G_m,...,G_z\},\\
    G_m=\{C_0,...,C_n\},\ (C_0 \cap \dots \cap C_n = 0). 
\end{gather} 
Here, $n$ is the number of semantic classes in this cluster $G_m$ ($n<k$).
Note that $n$ is not fixed to ensure adaptability across diverse tasks.
\lsy{Then, parameterization is implemented as the prior within each local cluster, $\alpha^{G_m}=[\alpha^{0},...,\alpha^{c_n}]$.}
\lsy{
\begin{equation}
    \alpha^{G_m} = \text{RELU}(D_d^k[G_m])+1,
\end{equation}
where $\text{RELU}$ is the active function. The evidence is $e^{G_m} = \text{RELU}(D_d^k[G_m])$. Similarly, predictive semantic probabilities and uncertainty measures are computed using the Dirichlet distribution. The difference is that uncertainty and probability are independently realized within each cluster $G_m$.}
\lsy{
\begin{equation}
    P_e^{G_m} = \alpha^{G_m} / S^{G_m}, \ \ \
    S^{G_m} = \sum_{j=0}^{C_n}\alpha^{j}.
\end{equation}
}

To train this head, the type-II maximum likelihood version of the uncertainty-aware loss is adopted in the local clusters:
\begin{equation}
\label{loss_edl}
    Loss_e^{G_m} = \sum_{i=0}^{H_b{\times}W_b}\sum_{j=0}^{C_n}L_i^{j} \cdot log(S/\alpha_i^{j}).
\end{equation}

Additionally, following the work of~\cite{evidential}, we introduce a KL divergence term $Loss_{kl}^{G_m}$ to enforce minimal evidence generation in non-labeled regions, ensuring the network only produces strong evidence for correct categories while suppressing others.
\begin{equation}
\label{klloss}
    Loss_{kl}^{G_m} = \text{KL}[D(P_e^k|\widetilde{\alpha_k}) || D(P_e^k|(1,\dots,1)) ],
\end{equation}
where $\widetilde{\alpha_k}$ represents the unlabeled region.
As described in Algorithm~\ref{code}, the training losses from all local clusters are aggregated to form the total loss for this segmentation head: 
\begin{equation}
    Loss_e = \sum_{G} (Loss_e^{G_m} + \lambda_k \times Loss_{kl}^{G_m}),
\end{equation}
where $\lambda_k=\text{min}(1,it/I)$ is controlled weight of KL divergence term. $it$ is the current iteration and $I$ is the training iteration.

\begin{table*}[t]
\fontsize{4}{5}\selectfont
\renewcommand{\arraystretch}{1.1}
\setlength\tabcolsep{7pt}
\caption{The unsupervised domain adaption performance (IoU\%) on different benchmarks of the nuScenes dataset~\cite{nus}. $*$ represents data from the work~\cite{pct}}
\vspace{-2.0em}
\label{tab:uda}
\begin{center}
\resizebox{0.9\linewidth}{!}{
\begin{tabular}{cccccccccc}
\toprule [1pt]
\multicolumn{1}{l|}{\multirow{2}{*}{Method}}  & \multicolumn{1}{l|}{\multirow{2}{*}{Image Size}} & \multicolumn{6}{c|}{IoU} & \multicolumn{1}{l}{\multirow{2}{*}{mIoU}} \\ \cline{3-8}
\multicolumn{1}{l|}{}   & \multicolumn{1}{l|}{} & Dri. & Ped. & Walk. & Stop. & Car. & \multicolumn{1}{l|}{Div.} & \multicolumn{1}{l}{}   \\ \hline
\multicolumn{9}{c}{\textbf{Boston $\xrightarrow{}$ Singapore}} \\ \hline
\multicolumn{1}{l|}{Source Only}  & \multicolumn{1}{l|}{\lsy{128${\times}$352}} & 40.2 & 6.3 & 7.0 & 1.9 & 0.7 & \multicolumn{1}{l|}{9.7} & \multicolumn{1}{l}{10.9}  \\
\multicolumn{1}{l|}{DualCross~\cite{man2023dualcross}}  & \multicolumn{1}{l|}{\lsy{128${\times}$352}} & 43.8  &8.2   &13.5   &4.5   &3.6    & \multicolumn{1}{l|}{9.6} & \multicolumn{1}{l}{13.9}  \\
\multicolumn{1}{l|}{DomainADV*~\cite{pct}} & \multicolumn{1}{l|}{\lsy{224${\times}$480}} & 40.0 & 8.3 & 11.7 & 4.5 & 2.2 & \multicolumn{1}{l|}{11.6} & \multicolumn{1}{l}{13.1}  \\
\multicolumn{1}{l|}{PCT*~\cite{pct}}  & \multicolumn{1}{l|}{\lsy{224${\times}$480}} & 46.2 & 8.6 & 14.2 & 6.4 & 3.7 & \multicolumn{1}{l|}{15.0} & \multicolumn{1}{l}{15.7}  \\
\multicolumn{1}{l|}{MT+PV}  & \multicolumn{1}{l|}{\lsy{128${\times}$352}} &49.3  &9.8   &12.9   &4.9   &4.5    & \multicolumn{1}{l|}{13.7} & \multicolumn{1}{l}{15.8} \\
\multicolumn{1}{l|}{Ours}  & \multicolumn{1}{l|}{\lsy{128${\times}$352}} & 59.1  &16.9   &21.9   &12.1   &16.8   & \multicolumn{1}{l|}{21.0 } & \multicolumn{1}{l}{\textbf{24.6}}  \\ \bottomrule [1pt]
\multicolumn{9}{c}{\textbf{Singapore $\xrightarrow{}$ Boston}} \\ \hline
\multicolumn{1}{l|}{Source Only}  & \multicolumn{1}{l|}{\lsy{128${\times}$352}} & 44.7 & 2.6 & 11.5 & 4.2 & 0.2 & \multicolumn{1}{l|}{8.4} & \multicolumn{1}{l}{11.9}  \\
\multicolumn{1}{l|}{DualCross~\cite{man2023dualcross}}  & \multicolumn{1}{l|}{\lsy{128${\times}$352}} & 45.7 &2.2 &13.6 &6.4 &5.2  & \multicolumn{1}{l|}{9.9} & \multicolumn{1}{l}{13.8}  \\
\multicolumn{1}{l|}{DomainADV*~\cite{pct}}  & \multicolumn{1}{l|}{\lsy{224${\times}$480}} & 35.7 & 4.2 & 11.3 & 4.8 & 0.6 & \multicolumn{1}{l|}{9.7} & \multicolumn{1}{l}{11.1}  \\
\multicolumn{1}{l|}{PCT*~\cite{pct}}  & \multicolumn{1}{l|}{\lsy{224${\times}$480}} & 47.0 & 8.0 & 19.3 & 6.3 & 0.7 & \multicolumn{1}{l|}{13.7} & \multicolumn{1}{l}{15.8} \\
\multicolumn{1}{l|}{MT+PV}  & \multicolumn{1}{l|}{\lsy{128${\times}$352}} & 48.0 &2.3 &16.8 &7.8 &4.9  & \multicolumn{1}{l|}{10.5} & \multicolumn{1}{l}{15.0} \\
\multicolumn{1}{l|}{Ours}  & \multicolumn{1}{l|}{\lsy{128${\times}$352}} & 61.4 &21.7 &31.8 &14.1 &24.1  & \multicolumn{1}{l|}{19.8} & \multicolumn{1}{l}{\textbf{28.8}} \\
 \bottomrule [1pt]
 \multicolumn{9}{c}{\textbf{Dry$\xrightarrow{}$ Rain}} \\ \hline
\multicolumn{1}{l|}{Source Only}  & \multicolumn{1}{l|}{\lsy{128${\times}$352}} & 67.1 & 29.5 & 35.8 & 23.4 & 24.6 & \multicolumn{1}{l|}{25.1} & \multicolumn{1}{l}{34.2}  \\
\multicolumn{1}{l|}{DualCross~\cite{man2023dualcross}}  & \multicolumn{1}{l|}{\lsy{128${\times}$352}} & 67.9 &29.8 &38.4 &22.6 &29.3 & \multicolumn{1}{l|}{25.4  } & \multicolumn{1}{l}{35.6}  \\
\multicolumn{1}{l|}{DomainADV*~\cite{pct}} & \multicolumn{1}{l|}{\lsy{224${\times}$480}} & 72.0 & 39.8 & 42.0 & 33.7 & 38.9 & \multicolumn{1}{l|}{33.6} & \multicolumn{1}{l}{43.3} \\
\multicolumn{1}{l|}{PCT*~\cite{pct}} & \multicolumn{1}{l|}{\lsy{224${\times}$480}} & 78.3 & 45.2 & 52.1 & 37.6 & 47.2 & \multicolumn{1}{l|}{36.4} & \multicolumn{1}{l}{\textbf{49.5}} \\
\multicolumn{1}{l|}{MT+PV}  & \multicolumn{1}{l|}{\lsy{128${\times}$352}} & 70.5 &34.1 &39.9 &27.1 &29.2  & \multicolumn{1}{l|}{27.8} & \multicolumn{1}{l}{38.1} \\
\multicolumn{1}{l|}{Ours}  & \multicolumn{1}{l|}{\lsy{128${\times}$352}} & 72.8 &36.6 &42.5 &29.8 &41.7  & \multicolumn{1}{l|}{30.4} & \multicolumn{1}{l}{42.3}  \\ \bottomrule [1pt]
\multicolumn{9}{c}{\textbf{Day $\xrightarrow{}$ Night}} \\ \hline
\multicolumn{1}{l|}{Source Only}  & \multicolumn{1}{l|}{\lsy{128${\times}$352}} & 32.8 & 2.2 & 4.3 & 4.4 & 0.0 & \multicolumn{1}{l|}{9.2} & \multicolumn{1}{l}{8.8}  \\
\multicolumn{1}{l|}{DualCross~\cite{man2023dualcross}}  & \multicolumn{1}{l|}{\lsy{128${\times}$352}} & 49.4 &8.0 &12.7 &7.9 &0.0   & \multicolumn{1}{l|}{15.1} & \multicolumn{1}{l}{15.5}  \\
\multicolumn{1}{l|}{DomainADV*~\cite{pct}}  & \multicolumn{1}{l|}{\lsy{224${\times}$480}} & 37.1 & 16.4 & 10.7 & 5.7 & 0.0 & \multicolumn{1}{l|}{11.2} & \multicolumn{1}{l}{15.1}  \\
\multicolumn{1}{l|}{PCT*~\cite{pct}} & \multicolumn{1}{l|}{\lsy{224${\times}$480}} & 51.3 & 19.4 & 16.1 & 7.6 & 0.0 & \multicolumn{1}{l|}{19.3} & \multicolumn{1}{l}{19.0}  \\
\multicolumn{1}{l|}{MT+PV}  & \multicolumn{1}{l|}{\lsy{128${\times}$352}} & 57.1 &9.6 &13.8 &6.2 &0.0 & \multicolumn{1}{l|}{18.9} & \multicolumn{1}{l}{17.6} \\
\multicolumn{1}{l|}{Ours}  & \multicolumn{1}{l|}{\lsy{128${\times}$352}} & 61.1 &31.3 &18.6 &12.0 &0.0  & \multicolumn{1}{l|}{22.3} & \multicolumn{1}{l}{\textbf{24.2}}  \\  \bottomrule [1pt]
\end{tabular}}
\end{center}
\vspace {-3em}
\end{table*}
\subsection{The Overall Loss}
The overall loss for UDA and SSL tasks is defined as $Loss = Loss_s  + Loss_t$.
\begin{equation}
    Loss_s = Loss_{gt} + \lambda_1{\times}Loss_{pv}^{s} + \lambda_2{\times}Loss_{tr}^{s}+ \lambda_3{\times}Loss_e^{s}, 
\end{equation}
\begin{equation}
    Loss_t = \beta(Loss_{pl}+\lambda_3{\times}Loss_{e}^{t}) + \lambda_1{\times}Loss_{pv}^{t}+ \lambda_2{\times}Loss_{tr}^{t},
\end{equation}
where $Loss_{gt}$ is the Dice loss with the PGCM module for the source domain. $Loss_{pl}$ and $Loss_{e}^{t}$ are the supervision of the pseudo-probabilistic labels of the target domain, constrained by the L2 loss. $\lambda_1=0.5$ and $\lambda_2=0.1$ are referenced from the works~\cite{streammapnet,hierdamap}.
The $\lambda_3$ is set as $1$.
$\beta$  is controlled by a sigmoid ramp-up function, which starts at $0$ and gradually increases to $0.1$ when the training round is halfway through.

\section{Experiment}

\begin{table*}[tb]
\fontsize{4}{5}\selectfont
\renewcommand{\arraystretch}{1.1}
\setlength\tabcolsep{7pt}
\caption{The semi-supervised learning performance (IoU\%) on different benchmarks of the nuScenes dataset~\cite{nus}. $*$ represents that only synthetic data of PerDiff is adopted for the training.}
\vspace{-2.0em}
\label{tab:ssl}
\begin{center}
\resizebox{0.9\linewidth}{!}{
\begin{tabular}{lcccccccc}
\toprule[1.0pt]
\multicolumn{1}{l|}{\multirow{2}{*}{Method}} &\multicolumn{1}{l|}{\multirow{2}{*}{backbone}}  & \multicolumn{6}{c|}{IoU}                                      & \multirow{2}{*}{mIoU} \\ \cline{3-8}
\multicolumn{1}{l|}{} &\multicolumn{1}{l|}{}         & Dri. & Ped. & Walk. & Sto. & Car. & \multicolumn{1}{l|}{Div.} &                       \\ \midrule[1.0pt]
\multicolumn{9}{c}{1/8}      \\ \hline
\multicolumn{1}{l|}{Sup. Only}  &\multicolumn{1}{l|}{EfficientNet-B0} &62.5  &27.4 &33.2 &19.3 &12.7  & \multicolumn{1}{l|}{21.8} & 29.5                \\
\multicolumn{1}{l|}{Ours}  &\multicolumn{1}{l|}{EfficientNet-B0}       &63.8 &27.9 &33.5 &19.5 &23.7 
   & \multicolumn{1}{l|}{22.7} & \textbf{31.9}                        \\ \midrule[1.0pt]
\multicolumn{9}{c}{1/4}                      \\ \hline
\multicolumn{1}{l|}{Sup. Only} &\multicolumn{1}{l|}{EfficientNet-B0} & 65.0 & 28.6 & 34.6  & 20.6 & 30.3 & \multicolumn{1}{l|}{25.3} & 34.0                  \\
\multicolumn{1}{l|}{PCT~\cite{pct}}  &\multicolumn{1}{l|}{EfficientNet-B4}      & \multicolumn{6}{c|}{--}                                       & 41.9                  \\
\multicolumn{1}{l|}{MT+PV}  &\multicolumn{1}{l|}{EfficientNet-B0}    & 65.9 & 30.2 & 36.9  & 22.1 & 27.5 & \multicolumn{1}{l|}{27.2} & 35.0                  \\
\multicolumn{1}{l|}{Ours}   &\multicolumn{1}{l|}{EfficientNet-B0}    &71.8      &42.0      &45.1       & 33.4     &50.6      & \multicolumn{1}{l|}{35.6}     &\textbf{46.4}                       \\ \midrule[1.0pt]
\multicolumn{9}{c}{1/2}                         \\ \hline
\multicolumn{1}{l|}{Sup. Only} &\multicolumn{1}{l|}{EfficientNet-B0} & 67.7 & 34.4 & 39.0  & 26.1 & 28.5 & \multicolumn{1}{l|}{29.6} & 37.6                  \\
\multicolumn{1}{l|}{PCT~\cite{pct}}  &\multicolumn{1}{l|}{EfficientNet-B4}     & \multicolumn{6}{c|}{--}                                       & \textbf{51.6}                 \\
\multicolumn{1}{l|}{MT+PV}  &\multicolumn{1}{l|}{EfficientNet-B0}  & 69.0 & 36.4 & 41.6  & 28.4 & 31.9 & \multicolumn{1}{l|}{31.1} & 39.7                  \\
\multicolumn{1}{l|}{Ours*}    &\multicolumn{1}{l|}{EfficientNet-B0}    &75.3      &45.4      &47.3       &37.1      &47.2      & \multicolumn{1}{l|}{38.0}     & 48.1                      \\ 
\multicolumn{1}{l|}{Ours}    &\multicolumn{1}{l|}{EfficientNet-B0}    &75.0      &48.3      &49.6       &39.6      &54.1      &\multicolumn{1}{l|}{40.2 }     & 51.1                     \\\bottomrule[1.0pt]
\end{tabular}}
\vspace {-3em}
\end{center}
\end{table*}

\subsection{Experiment Setup}
In this section, we verify the effectiveness of the proposed method on the public nuScenes dataset~\cite{nus}. 
The nuScenes dataset is officially partitioned into $750$ training scenes and $150$ validation scenes. 
Current world models follow this paradigm for training, leveraging BEV labels from the validation set to generate multi-view synthetic images.
Thus, all synthetic images in this study are generated from the semantic labels of the validation set. 
Each model produces approximately $6018$ frames of synthetic data.

The evaluation spans unsupervised domain adaptation and semi-supervised learning tasks on the nuScenes benchmark. 
For unsupervised domain adaptation tasks, we evaluate the effectiveness of the proposed method in both cross-region ( $Boston \xrightarrow{} Singapore$, $Singapore \xrightarrow{}Boston$) and cross-weather ($Day \xrightarrow{} Night$, and $Dry \xrightarrow{}Rain$) domain adaptation tasks. 
In each domain adaptation task, scenes serve as the fundamental unit, with all frames of a scene assigned to the same domain based on their metadata, following the DualCross~\cite{man2023dualcross} and PCT~\cite{pct}.
For the semi-supervised learning task, we conduct validation under $1/8$, $1/4$, and $1/2$ labeled data ratios. 
The officially partitioned training and validation sets will be used as the benchmark. For each scene in the training set, a consecutive sequence of frames is proportionally sampled as labeled data, with the rest remaining unlabeled,  following the setup of~\cite{semibev}.

\subsection{Implementation Details}
\textbf{Synthetic Data from World Model:}
Three representative world models with personalized BEV control modules are chosen for benchmarking: PerlDiff~\cite{zhang2024perldiff}, MagicDrive~\cite{gao2023magicdrive}, and BEVControl~\cite{bevcontrol}.
Additionally, we follow the official configuration and inference utilizing the public pre-trained weights. BEVControl strictly adheres to its original configuration, generating a single set of synthetic data, whereas PerlDiff and MagicDrive each produce three distinct datasets. 
The three synthetic datasets are generated under different text-based control conditions: one uses the officially provided control text, while the other two sets modify the textual descriptions to specify nighttime and rainy weather environments, respectively. 

In subsequent experiments, in addition to using data generated solely by individual world models, we also mix data from different world models. Specifically, our mixing strategy randomly selects each frame from any available world model. 
Since this study employs a temporal fusion strategy, which involves feature integration across sequential frames, using different data sources for different frames effectively enables the fusion of novel data.

\textbf{Training Settings:}
The unsupervised domain adaptation and semi-supervised learning frameworks in this section are both established on the Mean Teacher (MT) architecture~\cite{MT}. 
\lsy{
Here, the student model is jointly trained on source and target domain data with strong data augmentation. The strong data augmentation methods include resize, rotation, cropping, flipping, and colorjitter.
The teacher model is updated via momentum based on the weights of the student model.}
The learning momentum of the teacher model is $\alpha=0.99$. 

LSS~\cite{LSS} is chosen as the pipeline of the BEV semantic segmentation task, where the size of a perspective image is $(128,352)$. 
The BEV semantic segmentation includes $6$ static classes. 
The perception range of BEV space is $(-50,50)m$ and the solution is $0.5m$, which is similar to PCT~\cite{pct}.
We use the mean Intersection over Union (mIoU) as the main evaluation metric.
When batch is $12$, the learning rate is set to $3e^{-3}$. PolyLR optimizer is used for all tasks with a minimum learning rate of $1e^{-5}$.
All experiments are conducted on NVIDIA RTX A6000 GPUs.

\subsection{Performance Comparison}
\textbf{Unsupervised Domain Adaption:}
To verify the competitive performance of our model, we conducted comparative experiments with representative methods, including DualCross~\cite{man2023dualcross} (adversarial learning framework) and PCT~\cite{pct} (mean teacher framework). 
Additionally, the baseline of our implementation adopts the mean teacher architecture with the supervision of the perspective label across all domains (MT+PV).
As shown in Table~\ref{tab:uda},  the proposed method achieves state-of-the-art performance across most domains.
For cross-region adaptation, the synthetic data follows official text descriptions incorporating diverse weather and environmental variations. 
For specialized domain shifts (rainy/night conditions), the textual generation parameters were explicitly aligned with target domain environmental characteristics. 
Through the integration of synthetic data with the proposed PGCM and BiDPP modules, the framework achieves a maximum accuracy improvement of $13.8\%$ mIoU compared to baseline methods (MT+PV) at cross-domain $Singapore \xrightarrow{}Boston$.

\begin{table}[tb]
\fontsize{8.5}{9.2}\selectfont
\renewcommand{\arraystretch}{1.1}
\setlength\tabcolsep{7pt}
\caption{Ablation on the core modules. PGCM denotes the Perspective-Geometry Consistency Metric module. 
BiDPP denotes the Bi-Distribution Parallel Learning module.}
\vspace{-1.5em}
\label{tab:core}
\begin{center}
\resizebox{0.9\linewidth}{!}{
\begin{tabular}{lccccc}
\toprule[1.5pt]
\multicolumn{1}{l|}{Method}                       & Stream & Sys. Data & PGCM & \multicolumn{1}{l|}{BiDPP} & mIoU \\ \midrule[1.5pt]
\multicolumn{6}{c}{\textbf{Boston$\xrightarrow{}$ Singapore}}                                                                                           \\ \hline
\multicolumn{1}{l|}{\multirow{3}{*}{Sup. Only}} &        &           &      & \multicolumn{1}{l|}{}      & 10.9 \\
\multicolumn{1}{l|}{}                             & \checkmark      &           &      & \multicolumn{1}{l|}{}      & 13.8 \\
\multicolumn{1}{l|}{}                             & \checkmark       &\checkmark          &      & \multicolumn{1}{c|}{}      & 16.5 \\ \hline
\multicolumn{1}{c|}{\multirow{5}{*}{UDA}}         &        &           &      & \multicolumn{1}{l|}{}      & 15.8 \\
\multicolumn{1}{l|}{}                             & \checkmark       &           &      & \multicolumn{1}{l|}{}      & 18.6 \\
\multicolumn{1}{l|}{}                             & \checkmark       & \checkmark          &      & \multicolumn{1}{c|}{}      & 22.6 \\
\multicolumn{1}{l|}{}                             & \checkmark       & \checkmark          & \checkmark     & \multicolumn{1}{l|}{}      & 23.8 \\
\multicolumn{1}{l|}{}                             & \checkmark       & \checkmark          & \checkmark     & \multicolumn{1}{c|}{\checkmark }     & 24.6 \\ \midrule[1.5pt]
\multicolumn{6}{c}{\textbf{Day$\xrightarrow{}$ Night}}                                                                                          \\ \hline
\multicolumn{1}{c|}{\multirow{4}{*}{UDA}}         &        &           &      & \multicolumn{1}{c|}{}      & 17.6 \\
\multicolumn{1}{c|}{}                             & \checkmark       & \checkmark          &      & \multicolumn{1}{c|}{}      & 23.0 \\
\multicolumn{1}{c|}{}                             & \checkmark       & \checkmark         & \checkmark     & \multicolumn{1}{c|}{}      & 23.4 \\
\multicolumn{1}{c|}{}                             & \checkmark       & \checkmark         & \checkmark    & \multicolumn{1}{c|}{\checkmark}     & 24.2 \\ \midrule[1.5pt]
\multicolumn{6}{c}{\textbf{$1/8$}}                                                                                          \\ \hline
\multicolumn{1}{c|}{\multirow{3}{*}{SSL}}         &        &\checkmark      &      & \multicolumn{1}{c|}{}      & 30.8 \\
\multicolumn{1}{c|}{}                             &        &\checkmark        &\checkmark      & \multicolumn{1}{c|}{}      & 31.4 \\
\multicolumn{1}{c|}{}                             &       & \checkmark         & \checkmark     & \multicolumn{1}{c|}{\checkmark}      & 31.9 \\ \bottomrule[1.5pt]
\end{tabular}}
\end{center}
\vspace {-2.0em}
\end{table}
\begin{table}[tb]
\fontsize{9}{12}\selectfont
\renewcommand{\arraystretch}{1.1}
\setlength\tabcolsep{9pt}
\caption{Results of denoising experiments with different synthetic datasets generated via the driving world models, evaluated on the cross-domain setup of $Boston \xrightarrow{} Singapore$}.
\vspace{-1.0em}
\label{tab:denoise}
\begin{center}
\resizebox{0.9\linewidth}{!}{
\begin{tabular}{ccc|c|c}
\toprule[1.5pt]
Perl.~\cite{zhang2024perldiff} & Magic.~\cite{gao2023magicdrive} & BEVCon.~\cite{bevcontrol} & PGCM & mIoU \\ \midrule[1.0pt]
\lsy{FID(24.4)}  &\lsy{FID(25.8)}  &\lsy{FID(54.5)} & & \\ \hline
\checkmark     &      &      &      & 21.0 \\
\checkmark     &      &      & \checkmark    & 21.5 \\ \hline
      & \checkmark    &      &      & 20.7 \\
      & \checkmark    &      & \checkmark    & 21.1 \\\hline
      &      & \checkmark    &      & 20.0 \\
      &      & \checkmark    & \checkmark    & 20.5 \\\hline
\checkmark     & \checkmark    &      &      & 21.7 \\
\checkmark     & \checkmark    &      & \checkmark    & 22.8 \\ \bottomrule[1.5pt]
\end{tabular}}
\end{center}
\vspace{-1.0em}
\end{table}

\textbf{Semi-Supervised Learning:}
For semi-supervised learning benchmarks, we selected PCT~\cite{pct} and the standard MT with the perspective label as the baselines for comparisons. 
Our empirical observations reveal that model accuracy degrades when the volume of synthetic data exceeds that of the original labeled data. 
Since the labeled data proportions vary across tasks, the use of synthetic data is differentially configured for different tasks.
For our experiments, we employed PerlDiff~\cite{zhang2024perldiff} and MagicDrive~\cite{gao2023magicdrive} as the synthetic data generator due to their demonstrated higher-quality outputs compared to alternative approaches. 
Furthermore, in the $1/8$ semi-supervised learning setting, we randomly select $2,000$ frames of synthetic data from one world model as additional labeled data, ensuring that the newly introduced data volume remains smaller than the original labeled dataset. For the $1/4$ semi-supervised task, the full dataset generated by PerlDiff is adopted, while the $1/2$ task employs an aggregated dataset comprising both PerlDiff and MagicDrive outputs.
As evidenced in Table~\ref{tab:ssl}, the proposed learning paradigm significantly enhances model accuracy, demonstrating its effectiveness for semi-supervised learning tasks. 
Specifically, our method outstrips the previous state-of-the-art method PCT by $4.5\%$ in mIoU for the SSL of $1/4$ tasks.
But the proposed method exhibits marginally inferior performance in the $1/2$ labeled data SSL setting.
This phenomenon arises due to a more lightweight backbone, which has been proven to affect the accuracy of the BEV model~\cite{pct}.
Additionally, since the detailed architecture of the PCT model is unclear, we are unable to conduct an in-depth comparison.

\begin{figure*}[htb]
      \centering
      \includegraphics[scale=0.4]{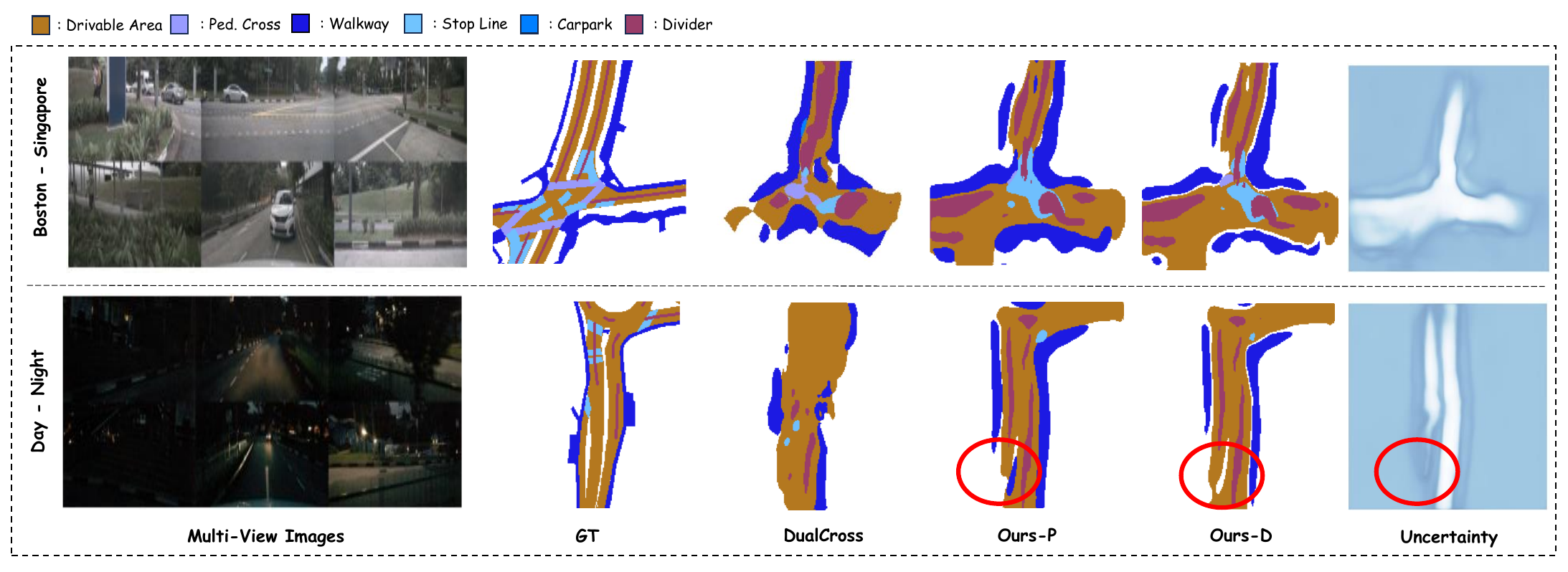}
      \vspace {-1.0em}
      \caption{Visualization results for unsupervised domain adaptation. It presents a visual comparison of our method against DualCross work~\cite{man2023dualcross}. Our method represents bi-distribution prediction results, `Ours-P' and `Ours-D', respectively. Additionally, we also present the uncertainty prediction results, where darker colors indicate higher uncertainty. It can be observed that our method has stronger cross-domain adaptation ability.}
      \label{fig.uda}
      \vspace {-1.0em}
\end{figure*}

\subsection{Ablation Studies}
In ablation studies, we first analyze the effectiveness of the proposed modules across different tasks. 
Then, we investigate the impact of labeled data quantity and quality on the performance of unsupervised learning. 
Finally, we conduct an in-depth analysis of the detailed configurations of the proposed modules for BEV segmentation.

\textbf{Ablation on the Core Modules:}
The baseline of the learning paradigm is the MT+PV framework. On this basis, the effectiveness of streaming temporal fusion, the PGCM module, and the BiDPP module is sequentially investigated.
As shown in Table~\ref{tab:core}, in the source-only training scenario, incorporating both temporal modeling and synthetic data yields a significant accuracy improvement of $5.6\%$ in mIoU, conclusively validating the efficacy of the temporal fusion module based on newly labeled data. 
In both UDA tasks, the introduction of labeled synthetic data significantly enhanced accuracy. Furthermore, the proposed PGCM and BiDPP modules demonstrate additional improvements in data utilization efficiency, delivering accuracy gains of $1.2\%$ and $0.8\%$ respectively in the setting of $Boston \xrightarrow{} Singapore$.
Similarly, in the semi-supervised learning task, embeddings from both modules effectively contributed to performance improvements, collectively achieving an accuracy gain of $1.1\%$ in mIoU.

\begin{figure}[tb]
      \centering
      \includegraphics[scale=0.3]{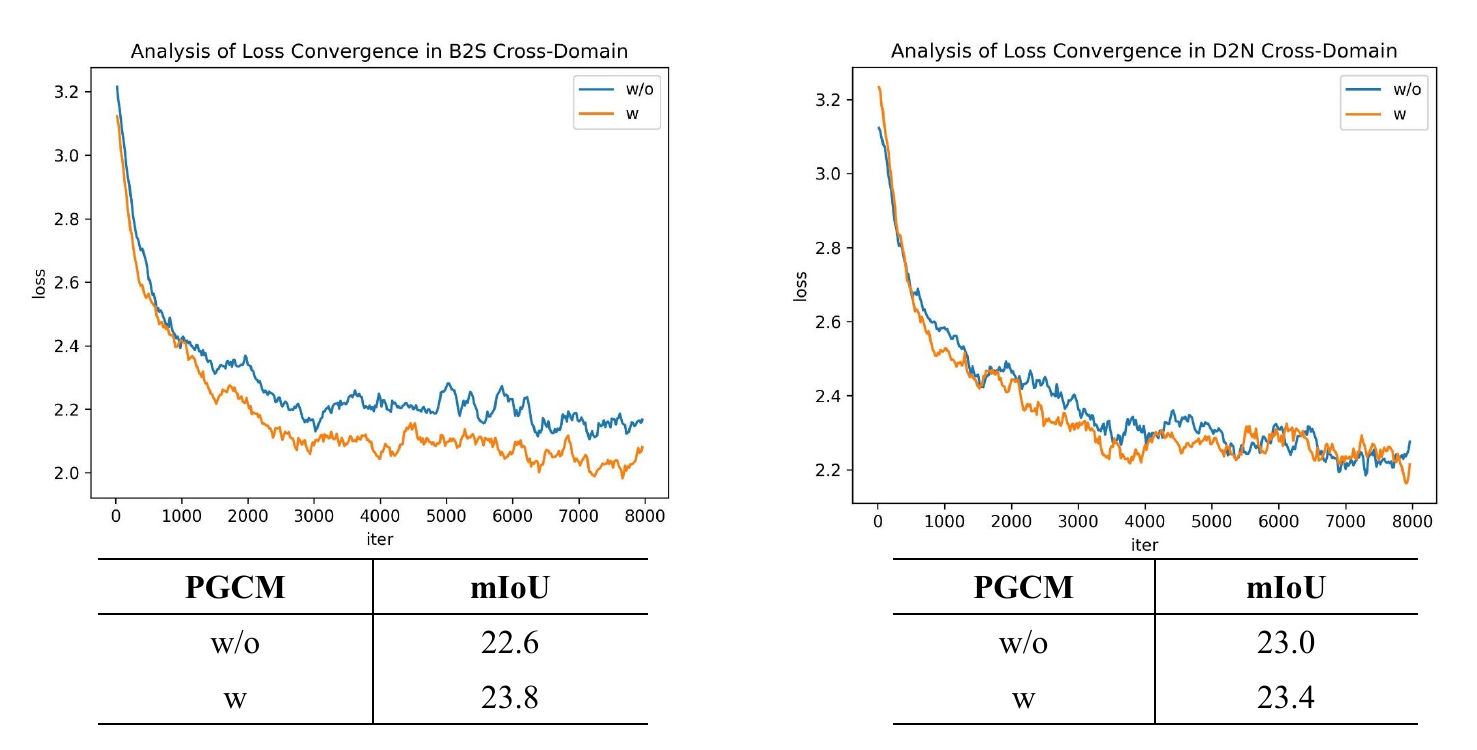}
      \caption{\lsy{Analysis of the convergence of training loss for the PGCM module. `w' indicates that the PGCM module is used, and `w/o' means not used.}}
      \label{fig.loss_ana}
      \vspace {-1.0em}
\end{figure}

\lsy{\textbf{Ablation on Denoising Effect of PGCM Module:}}
To verify the broad applicability of the proposed PGCM module, this subsection assesses its denoising effectiveness across synthetic data generated by different world models.
\lsy{The result is shown in Table~\ref{tab:denoise}. 
As observed from the FID scores, the synthetic data generated by PerfDiff~\cite{zhang2024perldiff} exhibits relatively high quality. 
Consequently, in the cross-domain BEV segmentation task, the synthetic data from this model contributes the most significant positive impact, achieving $21.0\%$ in mIoU.}
Additionally, the PGCM module demonstrates consistent effectiveness in improving BEV semantic segmentation accuracy, whether using synthetic data from individual models or combined data from multiple models.

\lsy{In addition, this section analyzes the convergence after loss optimization. As shown in Fig.~\ref{fig.loss_ana}, this section analyzes the loss convergence under different cross-domain settings. With the inclusion of the PGCM module, the model still demonstrates efficient convergence and achieves higher accuracy, with an increase of $1.2\%$ mIoU and $0.4\%$ mIoU, respectively.}

\begin{table}[tb]
\fontsize{10}{13.8}\selectfont
\renewcommand{\arraystretch}{1.1}
\setlength\tabcolsep{9pt}
\caption{Results with different loss optimization configurations. The configurations of the experiments are for the $1/8$ semi-supervised learning.}
\vspace{-1.2em}
\label{tab:loss}
\begin{center}
\resizebox{0.9\linewidth}{!}{
\begin{tabular}{c|cccccc|c}
\toprule[2pt]
Method                  & \multicolumn{6}{c|}{IoU}                & mIoU \\ \midrule[1.3pt]
Baseline  & 63.2 & 27.7 & 33.3 & 19.9 & 18.4 & 22.3 & 30.8 \\
M1        & 63.2 & 27.1 & 33.5 & 19.0 & 19.8 & 22.4 & 30.8 \\
M2        & 63.1 & 26.9 & 33.3 & 19.2 & 17.9 & 22.4 & 30.5 \\
Ours       & 63.3 & 27.9 & 33.7 & 19.4 & 21.6 & 22.1 & 31.4 \\ \bottomrule[2pt]
\end{tabular}}
\end{center}

\fontsize{6}{8}\selectfont
\renewcommand{\arraystretch}{1.1}
\setlength\tabcolsep{9pt}
\caption{\lsy{Results of different loss functions. The synthetic data from the PerfDiff~\cite{zhang2024perldiff} is selected as the baseline.}}
\vspace{-1.2em}
\label{tab:lossfunc}
\begin{center}
\resizebox{0.9\linewidth}{!}{
\begin{tabular}{l|cc|c}
\toprule[1.2pt]
Loss Function   & Sys. Data & PGCM  & mIoU \\ \midrule[0.7pt]
Focal Loss   & Perl.     &          & 18.7 \\ \hline
\multirow{2}{*}{DICE Loss}  &  Perl.      &           & 21.0 \\
    &  Perl.     & \checkmark         & 21.5 \\ 
 \bottomrule[1.2pt]
\end{tabular}}
\end{center}
\vspace{-1.5em}
\end{table}
\begin{table}[tb]
\fontsize{6}{8}\selectfont
\renewcommand{\arraystretch}{1.1}
\setlength\tabcolsep{9pt}
\caption{\lsy{Analysis on the loss weight of different distributions. $Loss_{gt}$ corresponds to the multinomial distribution loss, and $loss_e$ corresponds to the Dirichlet distribution loss.}}
\vspace{-1.2em}
\label{tab:wright}
\begin{center}
\resizebox{0.9\linewidth}{!}{
\begin{tabular}{cc|c|c}
\toprule[1.2pt]
$Loss_{gt}$   & $Loss_e$ & PGCM  & mIoU \\ \midrule[0.7pt]
 1   &  0.1      & \checkmark          & 23.7 \\ 
 1   &  0.5     & \checkmark         & 24.0 \\
 1   &  1.0   & \checkmark         & 24.6 \\ \bottomrule[1.2pt]
\end{tabular}}
\end{center}
\vspace{-1.5em}
\end{table}

\lsy{\textbf{Ablation on the Structure of PGCM Module:}}
This section first conducted comparative experiments with alternative loss optimization approaches.
Specifically, following the work~\cite{li2019dice-2}, we implemented two distinct optimization schemes.
\begin{equation}
    \text{M1} = 1-\frac{\textbf{R}*2*P_{+}}{\textbf{R}*(P_{+} + P_{-})+N_{+}+\sigma}.
\end{equation}
\begin{equation}
    \text{M2}  = 1-\frac{(2-\textbf{R})*2*P_{+}}{(2-\textbf{R})*(P_{+} + P_{-})+N_{+}+\sigma}.
\end{equation}
The primary objective of M1 is to employ a score $S$ to control the predicted probability $P_+$, preventing it from easily converging to $1$ during learning. In contrast, M2 utilizes $S$ to accelerate and guide the synthetic data to align with the labeled data direction.
The baseline of this section is built on the SSL of $1/8$ setting.
As shown in Table~\ref{tab:loss}, the results demonstrate that the former has a minimal impact on learning, whereas the latter hurts model performance. 
This negative effect stems from the overly efficient convergence of synthetic data, which leads to erroneous learning directions.
Our approach successfully rectifies these suboptimal learning directions and reaches a better performance.

\lsy{Subsequently, in this section, we further conduct experiments targeting different loss functions, where the Focal Loss is selected as the comparative loss function.
The result is shown in Table~\ref{tab:lossfunc}. It can be observed that the performance of Focal Loss is inferior to that of DICE loss when synthetic data is added to the source domain data.
This is because the Focal Loss strongly constrains learning at the individual pixel level, which can lead to erroneous fitting. In contrast, DICE loss exhibits higher tolerance for subtle structural misalignments. 
}

\lsy{\textbf{Ablation on the Loss Weight of BiDPP:}}
\lsy{
Both distributions in the BiDPP module are subject to label constraints. This section further analyzes the impact of the constraint loss weights.
As shown in Table~\ref{tab:wright}, it can be observed that when the weight of the multinomial distribution loss $Loss_{gt}$ is fixed at $1$, the performance is optimal when the weights of the Dirichlet distribution loss $Loss_{e}$ are relatively balanced. This configuration allows both to function synergistically.
}

\begin{table}[tb]
\fontsize{14}{15.8}\selectfont
\renewcommand{\arraystretch}{1.1}
\setlength\tabcolsep{9pt}
\caption{Analysis on the amount of synthetic data on UDA. `O-S' means the original size of the labeled data, whereas `N-S' represents the added labeled data.}
\vspace{-0.8em}
\label{tab:data_volum}
\begin{center}
\resizebox{0.9\linewidth}{!}{
\begin{tabular}{ccccccc}
\toprule[2pt]
Perl.~\cite{zhang2024perldiff} & Magic.~\cite{gao2023magicdrive} & BEVCon.~\cite{bevcontrol} & \multicolumn{1}{c|}{Mix} & \multicolumn{1}{c|}{O-S} & \multicolumn{1}{c|}{N-S} & mIoU \\ \midrule[2pt]
         &            &            &\multicolumn{1}{c|}{}        & \multicolumn{1}{c|}{\multirow{7}{*}{\lsy{18k}}}      &\multicolumn{1}{c|}{--}                      & 17.3 \\ \hline
\checkmark        &            &            &\multicolumn{1}{c|}{}        & \multicolumn{1}{c|}{}                           & \multicolumn{1}{c|}{\lsy{6k} }                 & 21.0 \\
         &\checkmark           &            & \multicolumn{1}{c|}{}       &\multicolumn{1}{c|}{}                            & \multicolumn{1}{c|}{\lsy{6k}}                  & 20.7 \\
         &            & \checkmark           & \multicolumn{1}{c|}{}       &\multicolumn{1}{c|}{}                            & \multicolumn{1}{c|}{\lsy{6k}}                  & 20.0 \\
\checkmark         & \checkmark           &            & \multicolumn{1}{c|}{}       &\multicolumn{1}{c|}{}                            & \multicolumn{1}{c|}{\lsy{12k}}                  & 21.7 \\
\checkmark         & \checkmark           &            & \multicolumn{1}{c|}{\checkmark}      &\multicolumn{1}{c|}{}                            & \multicolumn{1}{c|}{\lsy{18k} }                 & 22.6 \\
\checkmark         &\checkmark          & \checkmark          &\multicolumn{1}{c|}{\checkmark}       &  \multicolumn{1}{c|}{}                          & \multicolumn{1}{c|}{\lsy{24k}}                  & 22.0\\ \bottomrule[2pt]
\end{tabular}}
\end{center}
\vspace{-1.05em}
\end{table}
\textbf{Ablation on the Amount of Training Data:}
This paper aims to utilize diversified synthetic data to enhance the capabilities of BEV semantic segmentation models. In this subsection, we examine the impact on the amount of diversified synthetic data on BEV model learning.
As shown in Table~\ref{tab:data_volum}, when the amount of newly added synthetic data is less than the original labeled data, the improvement in accuracy is proportional to the amount of data. 
However, when the amount of synthetic data exceeds that of the original labeled data, the accuracy begins to decline. 
This occurs because the model tends to learn from the majority of the data, where the noisy synthetic data can misguide the learning.
Moreover, due to variations in image generation quality across different world models, it can be observed that the slightly inferior BEVControl~\cite{bevcontrol} provides only a limited improvement of $2.7\%$ in mIoU.
The more performant controlled PerlDiff~\cite{zhang2024perldiff} contributes a $3.7\%$ mIoU performance improvement.

\textbf{Ablation on Synthetic Image Quality:}
To further analyze the quality of synthetic data generation, we compared the learning performance between synthetic and real images in semi-supervised tasks.
Here, we divide the validation set into two parts; one part serves as newly labeled training data, while the other part continues to function as the validation set.
As shown in Table~\ref{tab:quality}, incorporating synthetic images as training data yielded inferior results compared to real images, indicating certain generative errors in the quality of synthetic data.
However, after optimizing the learning direction with the proposed PGCM module, the accuracy is improved significantly, demonstrating the effectiveness of PGCM.

\begin{table}[tb]
\fontsize{6}{7.8}\selectfont
\renewcommand{\arraystretch}{1.1}
\setlength\tabcolsep{9pt}
\caption{Ablation on the synthetic image quality. `Ori.' represents the original image data. `Perl.' denotes the synthetic data of the world model PerlDiff~\cite{zhang2024perldiff}.}
\vspace{-2.0em}
\label{tab:quality}
\begin{center}
\resizebox{0.9\linewidth}{!}{
\begin{tabular}{ccccc}
\toprule[1.2pt]
\multicolumn{1}{l|}{Data} & O-Size & \multicolumn{1}{l|}{N-Size} & \multicolumn{1}{l|}{PGCM} & mIoU \\ \midrule[1.2pt]
\multicolumn{5}{c}{SSL \ \ (1/8)} \\ \hline
\multicolumn{1}{l|}{Ori.} & \lsy{3k} & \multicolumn{1}{l|}{\lsy{2k}} & \multicolumn{1}{l|}{} & 31.6 \\
\multicolumn{1}{l|}{Perl.} & \lsy{3k} & \multicolumn{1}{l|}{\lsy{2k}} & \multicolumn{1}{l|}{} & 30.8 \\
\multicolumn{1}{l|}{Perl.} & \lsy{3k} & \multicolumn{1}{l|}{\lsy{2k}} & \multicolumn{1}{c|}{\checkmark} & 31.4 \\ \bottomrule[1.2pt]
\end{tabular}}
\end{center}
\vspace{-1.5em}
\end{table}

\textbf{Ablation on Different Text Settings:}
The world model can generate synthetic data under diverse environments by modifying text inputs, which proves highly beneficial for cross-domain adaptation learning. 
In our study of day-night cross-domain adaptation, we systematically compare the effects of different text configurations on domain transfer performance. As quantitatively demonstrated in Table~\ref{tab:weather}, compared to official text containing mixed weather conditions, synthetic data generated exclusively under nighttime settings further improves the accuracy by $1.2\%$ in mIoU.

\begin{table}[tb]
\fontsize{8}{10}\selectfont
\renewcommand{\arraystretch}{1.1}
\setlength\tabcolsep{9pt}
\caption{Ablation on different text conditions.}
\vspace{-1.2em}
\label{tab:weather}
\begin{center}
\resizebox{0.9\linewidth}{!}{
\begin{tabular}{l|cc|c}
\toprule[1.5pt]
Text                     & Stream & Sys. Data & mIoU \\ \midrule[1.0pt]
--                       &        &           & 17.6 \\
The official text        & \checkmark      & \checkmark         & 21.8 \\
The weather is nighttime & \checkmark      & \checkmark         & 23.0 \\ \bottomrule[1.5pt]
\end{tabular}}
\end{center}
\vspace{-1.5em}
\end{table}
\begin{table}[t]
\fontsize{7}{9}\selectfont
\renewcommand{\arraystretch}{1.1}
\setlength\tabcolsep{10pt}
\caption{Ablation on the method of perspective mask generation. `Sys. Mask' denotes the method to process synthetic data, whereas `Ref. Mask' denotes the reference mask. `GTProj' projects the BEV GT mask to the perspective view.}
\vspace{-1.0em}
\label{tab:masksel}
\begin{center}
\resizebox{0.9\linewidth}{!}{
\begin{tabular}{l|c|c|c}
\toprule[0.7pt]
Sys. Mask & Ref. Mask &\lsy{PGCM} & mIoU \\ \midrule[0.7pt]
                   &   &      & 21.7     \\ \hline
Mask2Former~\cite{mask2former}   &GTProj  &\lsy{\checkmark}     & 22.8    \\ 
SAN~\cite{san}                &GTProj     &\lsy{\checkmark}       & 22.0    \\ 
Mask2Former~\cite{mask2former}                 &Mask2Former~\cite{mask2former}     &\lsy{\checkmark}        & 22.9    \\ 
\bottomrule[0.7pt]                                 
\end{tabular}}
\end{center}
\vspace{-1.5em}
\end{table}
\begin{figure}[t]
      \centering
      \includegraphics[scale=0.45]{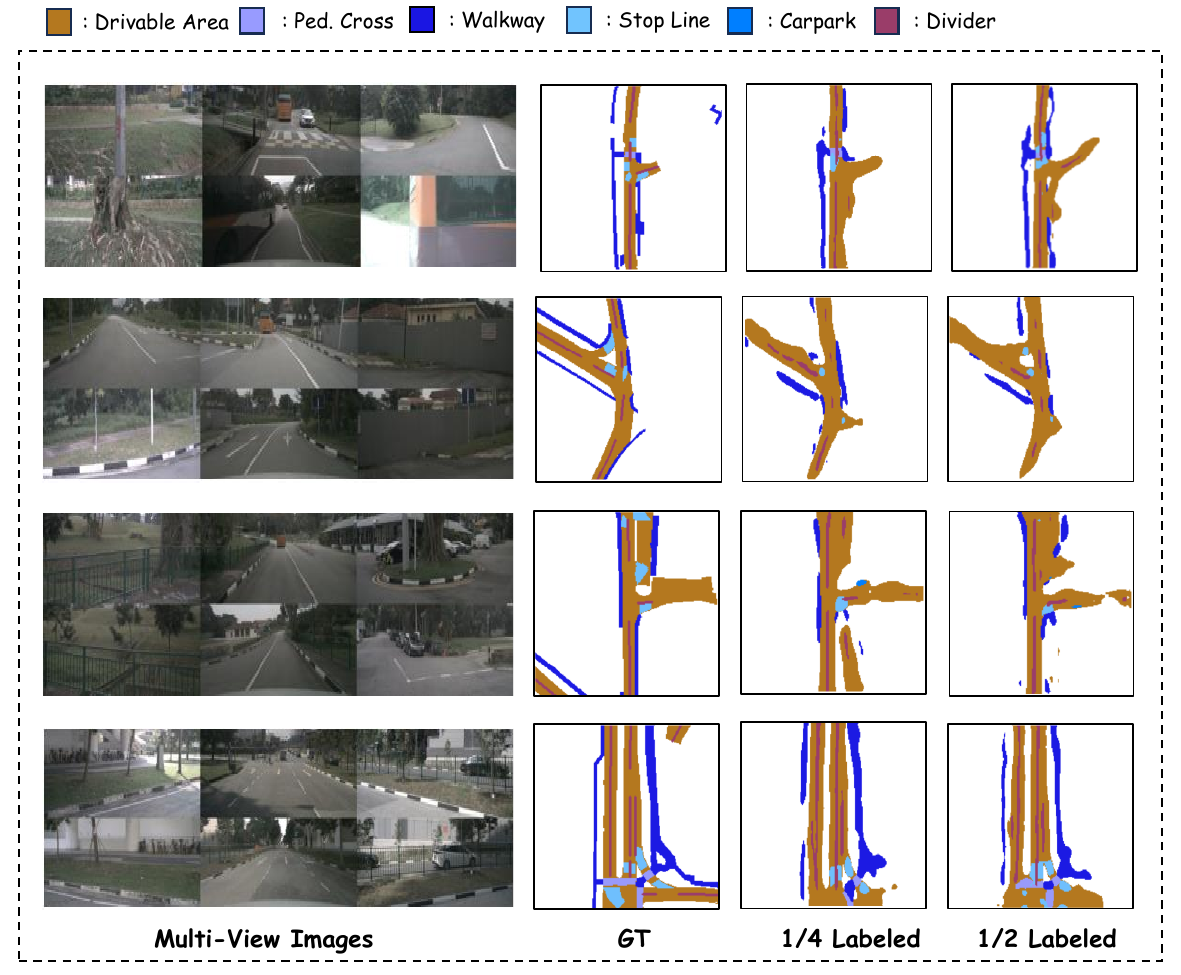}
      \vspace{-1.5em}
      \caption{Visualization results for semi-supervised learning. It shows the training results on 1/4 and 1/2 of the labeled data. It can be seen that the semantic results are more accurate for cases with more labeled samples.}
      \label{fig.ssl}
      \vspace {-1.5em}
\end{figure}
\textbf{Ablation on the Perspective Mask Method:}
The perspective consistency score is dependent on the accuracy of mask generation for synthetic data. Therefore, this section presents a comprehensive analysis of different mask models.
As shown in Table~\ref{tab:masksel}, we compare the performance between Mask2Former~\cite{mask2former} and SAN~\cite{san} models. 
The baseline is the UDA of $Boston\xrightarrow{} Singapore$ setting with the synthetic data from PerlDiff and MagicDrive, where PGCM does not work.
\lsy{Overall, the consistency scores derived from the masks generated by both pre-trained models can effectively mitigate noise issues and contribute to accuracy improvements.}
Additionally, the Mask2Former model is trained on Cityscapes, a dataset specifically designed for autonomous driving scenarios, which is similar to the nuScenes dataset. 
\lsy{The SAN model focuses on generalization capabilities under open-vocabulary conditions.
Therefore, our analysis indicates that the pre-trained Mask2Former model, which focuses on traffic data, generates more precise pseudo-labels.}
Furthermore, we considered another scenario where the original real images corresponding to the synthetic data exist. In this case, reference masks can also be generated using real images and a semantic segmentation model. The results show comparable performance to that obtained using BEV mask projection.

\begin{figure}[tb]
      \centering
      \includegraphics[scale=0.38]{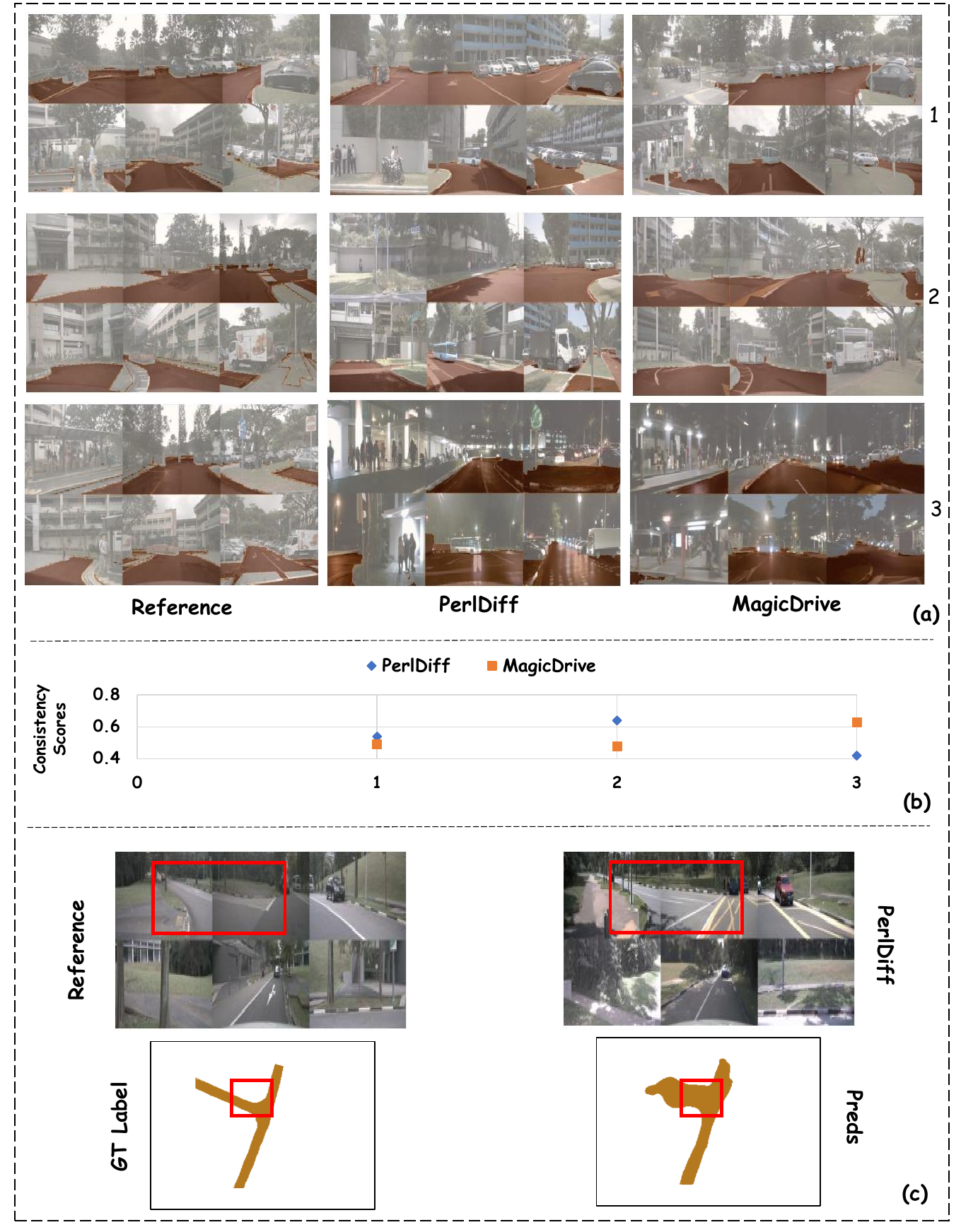}
      \caption{Visualization results and consistent scores of synthetic data from PerlDiff~\cite{zhang2024perldiff} and MagicDrive~\cite{gao2023magicdrive}. 
      (a) The first column displays the reference image of the dataset and the back-projected road mask. 
      The latter two columns present the generated synthetic data and predicted masks from Mask2Former~\cite{mask2former}.
      (b) Consistency scores $\textbf{R}$ of the synthetic data corresponding to the image examples from top to bottom.
      \lsy{(c) Compared to the reference image, the synthetic data within the red bounding box exhibits a slight offset in the road structure. Accordingly, the drivable area follows changed.}
      }
      \label{fig.world}
      \vspace {-1.0em}
\end{figure}

\subsection{Visualization Analyses}
\textbf{Unsupervised Domain Adaption:}
As illustrated in Fig.~\ref{fig.uda}, we compare the bi-distribution prediction results of our method with the DualCross approach~\cite{man2023dualcross} based on an adversarial learning framework.
Among them, `Ours-P' represents the results of probability prediction, whereas `Ours-D' denotes the low-uncertainty outcomes from the Dirichlet distribution. The latter considers reliable semantic predictions with uncertainty below $0.45$.
It can be observed that our method more accurately captures the road structure.
Furthermore, analyzing the results of bi-distribution prediction reveals that `Our-D' exhibits stronger capability in certain detailed predictions. As shown in the red-circled region of the nighttime semantic segmentation results, this area has relatively high uncertainty and thus does not belong to the walkway category.

\textbf{Semi-Supervised Learning:}
As shown in Fig.~\ref{fig.ssl}, the visualization results of semi-supervised learning with $1/4$ labeled samples and $1/2$ labeled samples are presented. 
The results demonstrate that the proposed method achieves outstanding performance in predicting BEV semantic segmentation. Additionally, it is evident that increasing the volume of labeled sample data significantly enhances the accuracy of BEV semantic segmentation.

\textbf{Synthetic Data from Different World Models:}
As shown in Fig.~\ref{fig.world}, we present synthetic data generated by different world models along with their consistency scores.
From an image generation quality perspective, the synthetic data demonstrates excellent quality.  The synthetic data aligns well with reference road masks in simple road environments. However, in complex environments such as curved lanes and parking lots, the generated images exhibit certain noise. 
Therefore, noise-resilient learning for synthetic data is essential to enhance BEV model training.

\lsy{
This paper designs a framework at both the data and architectural levels to prevent noisy synthetic data from overfitting to misaligned label data. As shown in Fig.~\ref{fig.world}(c), there are discrepancies in the road structure between the synthetic data and the reference image. The proposed method guides the model to learn the correct regions from the synthetic data.
}

\subsection{New-Split Fully-Supervised Performance}
The previous StreamMapNet~\cite{streammapnet} on vectorized high definition mapping introduced a novel dataset splitting strategy for nuScenes, specifically designed to evaluate BEV model generalization.
Compared to the original dataset splitting, the newly defined training and validation sets exhibit a significant distribution gap, leading to degraded generalization performance in BEV models.
To verify the effect of synthetic data on model generalization, we evaluate the proposed noise-resilient learning framework under fully supervised tasks on the newly split dataset.
As demonstrated in Table~\ref{tab:new_split}, the proposed NRSeg significantly enhances the accuracy of BEV semantic segmentation with an improvement of $3.3\%$ in mIoU, proving that incorporating diversely distributed data effectively improves the generalization capability.

\subsection{Cross-Dataset Adaptation Performance}
Due to the strong dependency of BEV models on training data characteristics, domain adaptation across datasets remains a significant challenge owing to inconsistencies in sensor configurations and data collection environments. Therefore, this section provides an analysis of our method for cross-dataset domain adaptation between Argoverse~\cite{argoverse} and nuScenes. 
Our experiments use Inverse Perspective Mapping (IPM) as the view transformation module, an approach that has been proven more suitable for cross-dataset BEV mapping~\cite{genmapping}.
As shown in Table~\ref{tab:crossdata}, after sequentially adding synthetic data from PerlDiff and MagicDrive, the cross-dataset BEV domain adaptation capability shows significant improvement.
The inclusion of synthetic data from a single world model yielded a $0.5\%$ precision improvement, while combining synthetic data from two world models delivered a $2.7\%$ precision boost.
This result further validates the effectiveness of our proposed noise-resilient learning framework for cross-dataset domain adaptation tasks.
\begin{table}[tb]
\fontsize{9}{13.8}\selectfont
\renewcommand{\arraystretch}{1.1}
\setlength\tabcolsep{9pt}
\caption{Results on the new-split nuScenes. It is implemented for the full-supervised BEV semantic segmentation.}
\vspace{-1.0em}
\label{tab:new_split}
\begin{center}
\resizebox{0.9\linewidth}{!}{
\begin{tabular}{l|cccccc|c}
\toprule[1.5pt]
\multirow{2}{*}{Method}  & \multicolumn{6}{c|}{IoU}                 & \multirow{2}{*}{mIoU} \\ \cline{2-7}
 & Dri. & Ped. & Walk. & Sto. & Car. & Div. &      \\\midrule[1.5pt]
LSS~\cite{LSS}    & 66.3     &29.6      &31.3       &16.1      & 19.0     &25.2      &31.3                       \\
NRSeg     &68.7      &31.5      &34.9       &19.5      &23.8      &29.0      &\textbf{34.6}                      \\ \bottomrule[1.5pt]
\end{tabular}}
\end{center}
\vspace {-1.5em}
\end{table}
\begin{table}[tb]
\fontsize{6.5}{7.5}\selectfont
\renewcommand{\arraystretch}{1.1}
\setlength\tabcolsep{9pt}
\caption{Results on cross-dataset domain adaptation. `VT' denotes the view transformer module.}
\vspace{-1.0em}
\label{tab:crossdata}
\begin{center}
\resizebox{0.9\linewidth}{!}{
\begin{tabular}{c|c|c|c}
\toprule [1pt]
Sys. Data & VT   & NRSeg & mIoU \\ \midrule[1pt]
          & LSS &       & 9.4  \\
          & IPM &       & 10.9 \\
PerlDiff~\cite{zhang2024perldiff}       & IPM &\checkmark       &\textbf{11.4}    \\ 
PerlDiff~\cite{zhang2024perldiff}+MagicDrive~\cite{gao2023magicdrive}      & IPM &\checkmark       &\textbf{13.6}    \\ 
\bottomrule[1pt]
\end{tabular}}
\end{center}
\vspace {-2.0em}
\end{table}

\section{Conclusion}
In this work, we found that incorporating diversely distributed synthetic data enhances BEV semantic segmentation performance, but its inherent noise degrades learning efficacy. 
To address this, we have proposed NRSeg, a noise-resilient learning framework for BEV semantic segmentation that optimizes the utilization of diverse and noisy synthetic data generated from driving world models. 
Specifically, we focus on achieving this through optimizing the guidance of noisy data and enhancing the inherent robustness of the model.
We conducted comprehensive evaluations on unsupervised domain adaptation and semi-supervised learning tasks, demonstrating that the proposed method achieves state-of-the-art performance. 
Simultaneously, for the newly partitioned nuScenes dataset, where the significant distribution gap between the training and validation sets adversely affects the generalization of models, we have verified that the proposed method effectively enhances the generalization capability of BEV semantic segmentation.
Additionally, we have demonstrated its capabilities in more challenging cross-dataset scenarios.

\lsy{\textbf{Limitations:} The proposed denoising framework effectively mitigates the inherent noise in such synthetic data. However, this framework requires co-training on the original source-domain data. When the volume of available source domain data is limited, the utility of noise synthetic data remains constrained.
}

\lsy{\textbf{Future Work:} In scenarios where only a pre-trained model on the source domain is available, approaches such as meta-learning can be employed to learn common task features from noisy synthetic data, thereby guiding learning on the unlabeled target domain.}

\bibliographystyle{IEEEtran}
\bibliography{refer.bib}

\end{document}